%% file: main.tex
\documentclass[runningheads]{llncs}

\usepackage{eccv}

\usepackage{eccvabbrv}

\usepackage{url}
\usepackage{marvosym}
\usepackage{graphicx}
\usepackage{booktabs}
\usepackage{multirow}
\usepackage{amssymb}
\usepackage{xcolor}
\usepackage{listings}
\usepackage{wrapfig}
\usepackage{ulem}

\usepackage[accsupp]{axessibility}  % Improves PDF readability for those with disabilities.

\newcommand{\diff}[1]{{\color{black}#1}}
\usepackage{tabularx} % 用于表格自动列宽
\usepackage{makecell} % 用于在单元格内分行
\usepackage{booktabs} % 提供了 \toprule, \midrule 和 \bottomrule
\usepackage{hyperref}

\usepackage{orcidlink}

\begin{document}

\title{Octopus: Embodied Vision-Language Programmer from Environmental Feedback} 

\titlerunning{Octopus: Embodied VLM from Environmental Feedback}

% Second Author\inst{2,3}\orcidlink{1111-2222-3333-4444} \and
% Third Author\inst{3}\orcidlink{2222--3333-4444-5555}}

\author{Jingkang Yang$^{*,1}$\and Yuhao Dong$^{*,2,3}$\and Shuai Liu$^{*,2,4}$\and Bo Li$^{*,1}$\and \\ Ziyue Wang$^{\dagger,1}$\and Haoran Tan$^{\dagger,4}$\and Chencheng Jiang$^{\dagger,5}$ \and Jiamu Kang$^{\dagger,3}$\and \\Yuanhan Zhang$^{1}$\and Kaiyang Zhou$^{6}$\and and Ziwei Liu$^{1,\text{\Letter}}$}
\authorrunning{J.~Yang et al.}

\institute{
$^{1}$ S-Lab, Nanyang Technological University \qquad
$^{2}$ Shanghai AI Laboratory\\
$^{3}$ Tsinghua University \qquad
$^{4}$ BUPT\qquad
$^{5}$ XJTU \qquad
$^{6}$ Hong Kong Baptist University \qquad
\email{\{jingkang001, ziwei.liu\}@ntu.edu.sg}
}

\maketitle
\renewcommand{\thefootnote}{\fnsymbol{footnote}}
\footnotetext[1]{Equal contribution, \textsuperscript{$\dagger$}Equal engineering contribution, \textsuperscript{\Letter} Corresponding author.}
\renewcommand*{\thefootnote}{\arabic{footnote}}
\begin{figure*}[h]
\vspace{-27pt}
\centering
    \includegraphics[width=0.99\linewidth]{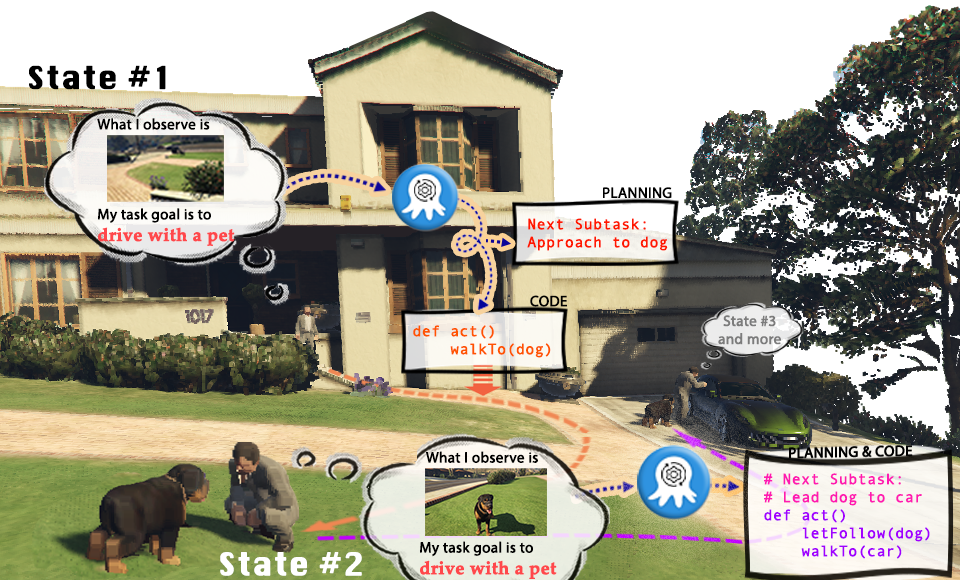}\\
    \vspace{-5pt}
    \caption{\small \textbf{Illustration of our vision-language programmer, Octopus, complete a task in GTA environment.}
    Given a task in the form of natural language, Octopus relies on its egocentric vision to generate plans and the corresponding executable code.
    }
    \vspace{-25pt}
    \label{fig:fig_teaser}
\end{figure*}

\begin{abstract}
Large vision-language models (VLMs) have achieved substantial progress in multimodal perception and reasoning. When integrated into an embodied agent, existing embodied VLM works either output detailed action sequences at the manipulation level or only provide plans at an abstract level, leaving a gap between high-level planning and real-world manipulation. To bridge this gap, we introduce \textbf{Octopus}, \textit{an embodied vision-language programmer} that uses executable code generation as a medium to connect planning and manipulation. Octopus is designed to \textbf{1)} proficiently comprehend an agent's visual and textual task objectives, \textbf{2)} formulate intricate action sequences, and \textbf{3)} generate executable code. 
To facilitate Octopus model development, we introduce \textbf{OctoVerse}: a suite of environments tailored for benchmarking vision-based code generators on a wide spectrum of tasks, ranging from mundane daily chores in simulators to sophisticated interactions in complex video games such as Grand Theft Auto (GTA) and Minecraft.
To train Octopus, we leverage GPT-4 to control an explorative agent that generates training data, i.e., action blueprints and corresponding executable code. We also collect feedback that enables an enhanced training scheme called \textbf{Reinforcement Learning with Environmental Feedback~(RLEF)}. Through a series of experiments, we demonstrate Octopus's functionality and present compelling results, showing that the proposed RLEF refines the agent's decision-making. By open-sourcing our simulation environments, dataset, and model architecture, we aspire to ignite further innovation and foster collaborative applications within the broader embodied AI community. The project page is available at \url{https://choiszt.github.io/Octopus/}.
\end{abstract}

\section{Introduction}
The rise of large language models (LLMs)~\cite{radford2019language, brown2020language, ouyang2022training,touvron2023llama,chiang2023vicuna} led to a surge in vision-language models (VLMs)~\cite{alayrac2022flamingo,awadalla2023openflamingo,li2023blip,li2023otter,liu2023llava,liu2023improvedllava,liu2024llavanext}, enabling tasks such as image/video-based descriptions~\cite{li2023blip}, reasoning~\cite{xie2023funqa,chen2023language,liu2024chain}, and conversations~\cite{dai2023instructblip,li2023otter}. In the realm of embodied AI, notable efforts~\cite{li2023blip, palme, rt2}
% like SayCan~\cite{saycan}, Palm-E~\cite{palme}, and RT-2~\cite{rt2} 
have trained agents to process visual input and relay motor control commands.

Another approach to interacting with the environment focuses on task execution through code invocations, mirroring the human System-I stimulation~\cite{evans2008dual,kahneman2011thinking} (automatic, intuitive actions) with predefined code, and leaving the System-II processes~\cite{evans2008dual,kahneman2011thinking} (planning and reasoning) for large models.
For example, referring to Fig.~\ref{fig:fig_teaser}, planning a car ride with a pet might entail a subconscious checklist (e.g., \texttt{getOutOf()} the house, \texttt{open()} the car door), each action could be implemented using specific techniques~\cite{billard2019trends, gu2017deep} such as imitation learning~\cite{hussein2017imitation, fu2024mobile}.
This programmatic paradigm has been, although not in vision, leveraged by works~\cite{schick2023toolformer,shen2023hugginggpt,suris2023vipergpt,gupta2023visual} 
% like ToolFormer~\cite{schick2023toolformer}, HuggingGPT~\cite{shen2023hugginggpt}, ViperGPT~\cite{suris2023vipergpt}, and VisProg~\cite{gupta2023visual}, 
using LLMs to craft programs and trigger APIs. Game-centric models like Voyager~\cite{wang2023voyager} have similarly employed GPT for function calls within game engines, though they often parse data directly from their environments.

However, when incorporating visual perception, the programming paradigms are largely unexplored. Primary initiatives~\cite{tapa,rana2023sayplan} can only output plans, which anchor their strategies only in initial environmental states or employ dynamic scene graphs for LLM inputs, respectively. Despite their innovations, the over-reliance on pre-trained vision models to convert vision content into language can occasionally hinder the LLM's planning performance. The conversion from plans into real-world actions is still missing. While EmbodiedGPT~\cite{embodiedgpt} addresses the problem by integrating vision-language modeling for planning and then transitioning to manipulation using policy mapping, the capability of embodied vision-language models to generate executable programs remains largely uncharted.

Our exploration aims to bridge this gap. An embodied vision-language programmer should integrate visual perspective with textual objectives to devise action plans and executable code (Fig.~\ref{fig:fig_teaser}). However, existing simulator environments often lack the carefully designed functions necessary to support such models effectively. These functions should balance usefulness and complexity to avoid hindering the development of genuine embodied vision-language programmers. For example, the \texttt{explore\_until()} function in Minecraft, which can lead the player directly to specific blocks without relying on vision information, may not be suitable for training these models.

To meet the requirement, we carefully design and develop OctoVerse, a suite of environments consisting of diverse simulators, including (i) OctoGibson, built upon the photorealistic OmniGibson~\cite{li2023behavior}, (ii)  OctoMC, developed on the infinitely creative, pixel-style Minecraft platform~\cite{Minecraft}, and (iii) OctoGTA, adapted from the highly interactive and immersive Grand Theft Auto V (GTA-V)~\cite{gtav2014}. These environments enable the training and benchmarking of our embodied vision-language programming model in a wide range of scenarios, from daily household tasks to complex urban navigation and open-world exploration, while the function calls are tailored to be vision-dependent.

Using the OctoVerse environment, we train Octopus by leveraging GPT-4 to collect data. We provide GPT-4 with system messages, environmental cues, and objectives, enabling it to formulate action strategies and code. Simultaneously, the agent captures visual perspectives, forming the image-code pair for Octopus training. During data collection, the agent receives simulator feedback, distinguishing successful moves from unsuccessful ones. We incorporate this feedback using Reinforcement Learning with Environmental Feedback (RLEF) and fine-tune Octopus using Proximal Policy Optimization (PPO)~\cite{schulman2017proximal}. Empirically, Octopus demonstrates strong adaptability in various scenarios, outperforming existing models in task planning, code generation, and execution. The integration of RLEF further enhances Octopus's performance, showcasing the effectiveness of this training approach. In sum, our key contributions include:
\begin{itemize}
    \item \textbf{A Novel Vision-Language Programming Benchmark:} Three diverse embodied environments with designed tasks: (i) OctoGibson, which is developed upon OmniGibson~\cite{li2023behavior}, (ii) OctoMC that developed on Minecraft~\cite{Minecraft}, and (iii) OctoGTA, which is adapted from GTA-V~\cite{gtav2014}.
    \item \textbf{A New Vision-Language Programming Model:} An embodied vision-language planner and programmer trained with Reinforcement Learning with Environmental Feedback (RLEF), demonstrates compelling results.
    \item \textbf{Insights on Vision-Language Programming:} We extensively explore Octopus and share useful insights facilitating future research on visual planning and programming.
\end{itemize}

\section{Related Work}

\subsection{Embodied AI Simulators}
Embodied AI has advanced significantly with the development of diverse simulation environments, enabling research tasks such as visual exploration~\cite{ramakrishnan2020exploration}, navigation~\cite{wang2022navigation}, and question-answering~\cite{das2017embodiedQA}. Several simulators, including AI2-THOR~\cite{kolve2017ai2}, VirtualHome~\cite{puig2018virtualhome}, Habitat-Sim~\cite{savva2019habitat}, SAPIEN~\cite{xiang2020sapien}, and Omnigibson~\cite{li2023behavior}, provide realistic representations of the world for investigating embodied AI challenges. OmniGibson~\cite{li2023behavior} stands out for its high-fidelity simulation of diverse indoor and outdoor environments. OctoGibson environment further enhances OmniGibson with carefully designed function calls and formulated tasks, making it well-suited for vision-language programming.

Game-related simulators like Arade~\cite{arcade}, CHALET~\cite{chalet}, and VRKitchen~\cite{vrkitchen} also contribute significantly to embodied AI. Minecraft~\cite{Minecraft} has gained attention in reinforcement learning and game agents~\cite{mao2021seihai,vpt,fan2022minedojo,plan4mc,wang2023voyager,zheng2023steveeye} but lacks the necessary structure for vision-language programming. OctoMC addresses this by providing designed function calls and formulated tasks. In contrast to Minecraft's voxel-based representations that limit transferability to real-world environments, GTA-V~\cite{gtav2014} offers a highly realistic environment. In this work, we introduce OctoGTA as a new setting, leveraging GTA-V's rich, open-world environment with incorporated tasks and function calls, extending this platform for embodied AI study.

\begin{table}[t]
\centering
\caption{\textbf{Related Work for OctoVerse - Overview of Embodied AI Environments.} 
We select three environments into OctoVerse and carefully design executable tasks and vision-dependent function calls (VC), in comparison to undesigned standard function calls (C).}
\label{tab:simulator}
\vspace{-8pt}
\resizebox{\textwidth}{!}{
\begin{tabular}{@{}lcccccccccc@{}}
\toprule
\textbf{Simulator} &
  \textbf{\begin{tabular}[c]{@{}c@{}}Kinematics\end{tabular}} &
  \textbf{\begin{tabular}[c]{@{}c@{}}Continuous\\Extended\\States\end{tabular}} &
  \textbf{\begin{tabular}[c]{@{}c@{}}Flexible\\Materials \end{tabular}} &
  \textbf{\begin{tabular}[c]{@{}c@{}}Deformable\\Bodies\end{tabular}}&
    \textbf{\begin{tabular}[c]{@{}c@{}}Realistic\\Action\\Execution \end{tabular}}&
    \textbf{\begin{tabular}[c]{@{}c@{}}Game- or\\ World-Based \end{tabular}}&
    \textbf{\begin{tabular}[c]{@{}c@{}}Formulated\\ Tasks\end{tabular}}&
    \textbf{\begin{tabular}[c]{@{}c@{}}Function\\Call\\Type \end{tabular}}
    \\
    \midrule
OpenAIGym~\cite{brockman2016openai}      & \textcolor{teal}{\checkmark} & \textcolor{red}{$\times$}  & \textcolor{red}{$\times$}&\textcolor{red}{$\times$}&
\textcolor{teal}{\checkmark}&
G&
\textcolor{red}{$\times$}&
\textcolor{teal}{C}
\\
Matterport3D~\cite{chang2017matterport3d} & \textcolor{red}{$\times$} & \textcolor{red}{$\times$} &
\textcolor{red}{$\times$} &
\textcolor{red}{$\times$} &
\textcolor{red}{$\times$}&
W&
\textcolor{red}{$\times$}&
\textcolor{red}{$\times$}
\\
AI2THOR~\cite{kolve2017ai2}               & \textcolor{teal}{\checkmark} & \textcolor{red}{$\times$}  & \textcolor{red}{$\times$} & \textcolor{red}{$\times$}& 
\textcolor{red}{$\times$}&
G&
\textcolor{red}{$\times$}&
\textcolor{teal}{C}
\\
VirtualHome~\cite{puig2018virtualhome}               & \textcolor{red}{$\times$} & \textcolor{red}{$\times$}  & \textcolor{red}{$\times$}& \textcolor{red}{$\times$}& 
\textcolor{red}{$\times$}&
G&
\textcolor{red}{$\times$}&
\textcolor{red}{$\times$}
\\
House3D~\cite{wu2018building}  
&\textcolor{red}{$\times$} & \textcolor{red}{$\times$}  & \textcolor{red}{$\times$}& \textcolor{red}{$\times$}& 
\textcolor{red}{$\times$}&
W&
\textcolor{red}{$\times$}&
\textcolor{red}{$\times$}
\\
Habitat 1.0~\cite{savva2019habitat}              & \textcolor{teal}{\checkmark} & \textcolor{red}{$\times$}  & \textcolor{red}{$\times$}& \textcolor{red}{$\times$}& 
\textcolor{teal}{\checkmark}&
W&
\textcolor{red}{$\times$}&
\textcolor{teal}{C}
\\
Robosuite~\cite{zhu2020robosuite}
& \textcolor{teal}{\checkmark} & \textcolor{red}{$\times$}  & \textcolor{red}{$\times$}& \textcolor{red}{$\times$}& 
\textcolor{teal}{\checkmark}&
W&
\textcolor{red}{$\times$}&
\textcolor{teal}{C}
\\
RFUniverse~\cite{fu2023rfuniverse}  & \textcolor{teal}{\checkmark} & \textcolor{red}{$\times$}  & \textcolor{teal}{\checkmark} & \textcolor{teal}{\checkmark}&
\textcolor{teal}{\checkmark}&
W&
\textcolor{red}{$\times$}  &
\textcolor{teal}{C}  
\\
\midrule
Minecraft~\cite{Minecraft}  & \textcolor{teal}{\checkmark} & \textcolor{red}{$\times$}  & 
\textcolor{teal}{\checkmark} & \textcolor{red}{$\times$}&
\textcolor{teal}{\checkmark}&
G&
\textcolor{red}{$\times$}    &
\textcolor{teal}{C}
\\
\textbf{OctoMC}  & \textcolor{teal}{\checkmark} & \textcolor{red}{$\times$}  & 
\textcolor{teal}{\checkmark} & \textcolor{red}{$\times$}&
\textcolor{teal}{\checkmark} &
G&
\textcolor{teal}{\checkmark}  &
\textcolor{teal}{VC}
\\
GTA~\cite{gtav2014}      & \textcolor{teal}{\checkmark} & \textcolor{teal}{\checkmark} & \textcolor{teal}{\checkmark} & \textcolor{teal}{\checkmark}&
\textcolor{teal}{\checkmark}&
G&
\textcolor{red}{$\times$}  &
\textcolor{teal}{C}
\\ 
\textbf{OctoGTA}       & \textcolor{teal}{\checkmark} & \textcolor{teal}{\checkmark} & \textcolor{teal}{\checkmark} & \textcolor{teal}{\checkmark}&
\textcolor{teal}{\checkmark}&
G&
\textcolor{teal}{\checkmark}&
\textcolor{teal}{VC}
\\ 
OmniGibson~\cite{li2023behavior}       & \textcolor{teal}{\checkmark} & \textcolor{teal}{\checkmark} & \textcolor{teal}{\checkmark} & \textcolor{teal}{\checkmark}&
\textcolor{teal}{\checkmark}&
W&
\textcolor{red}{$\times$}  &
\textcolor{teal}{C}
\\ 
\textbf{OctoGibson}       & \textcolor{teal}{\checkmark} & \textcolor{teal}{\checkmark} & \textcolor{teal}{\checkmark} & \textcolor{teal}{\checkmark}&
\textcolor{teal}{\checkmark}&
W&
\textcolor{teal}{\checkmark}&
\textcolor{teal}{VC}
\\
\bottomrule
\end{tabular}}
\vspace{-20pt}
\end{table}

\subsection{Embodied AI with Large Models}
The recent wave of research focuses on merging LLMs with embodied AI tasks~\cite{radford2019language, brown2020language, ouyang2022training,touvron2023llama}. For instance, VoxPoser addresses robotic manipulation problems through unsupervised methods~\cite{voxposer}. A group of projects, namely SayCan~\cite{saycan}, Palm-E~\cite{palme}, RT-2~\cite{rt2}, and EmbodiedGPT~\cite{embodiedgpt}, effectively integrate visual or linguistic cues with robot manipulation data. Outside the domain of robotic manipulation, initiatives like Voyager~\cite{wang2023voyager} and Smallville~\cite{park2023generative} harness the capabilities of GPT to interface with game functions, relying on preset functions to manage intricate manipulations. In a parallel vein, VisProg~\cite{gupta2023visual} leverages GPT-3 language prompts to craft Python programs, opening the door to a multitude of fascinating applications. While the proposed Octopus model also formulates plans and code, its distinguishing feature is the seamless integration of visual input in program and code generation. This also stands in contrast to other embodied planners like TAPA~\cite{tapa} and SayPlan~\cite{rana2023sayplan}, which deploy separate vision modules to translate visual data into linguistic inputs for LLMs. Octopus excels as a cohesive vision-language model, delivering not just plans but also code.

\begin{table}[t]
\centering
\caption{\textbf{Related Work for Octopus - Overviewing Embodied AI Models.} The proposed Octopus distinguishes itself from other models as a unified vision-language model for both plan and code generation.}
\vspace{-5pt}
\resizebox{\textwidth}{!}{
\begin{tabular}{@{}lcccccc@{}}
\toprule
\textbf{Models} &
  \textbf{\begin{tabular}[c]{@{}c@{}}Release\\ Date\end{tabular}} &
  \textbf{\begin{tabular}[c]{@{}c@{}}Supported\\ Environment\end{tabular}} &
  \textbf{\begin{tabular}[c]{@{}c@{}}Vision\\ Model\end{tabular}} &
  \textbf{\begin{tabular}[c]{@{}c@{}}Code\\ Generator\end{tabular}} &
  \textbf{\begin{tabular}[c]{@{}c@{}}Action\\ w/ Feedback\end{tabular}} &
  \textbf{\begin{tabular}[c]{@{}c@{}}LLM Training\\ Enabled\end{tabular}} \\ \midrule
Text2Motion~\cite{lin2023text2motion}  & Mar. 2023 & Sim & \textcolor{red}{$\times$} & \textcolor{teal}{\checkmark}  & \textcolor{teal}{\checkmark}  & \textcolor{red}{$\times$}  \\
Instruct2Act~\cite{huang2023instruct2act}  & May 2023 & Sim & \textcolor{red}{$\times$} & \textcolor{teal}{\checkmark}  & \textcolor{red}{$\times$}  & \textcolor{red}{$\times$}  \\
Lang2Rewards~\cite{yu2023language2rewards}  & Jun. 2023 & Sim & \textcolor{red}{$\times$} & \textcolor{teal}{\checkmark}  & \textcolor{teal}{\checkmark}  & \textcolor{red}{$\times$}  \\
VoxPoser~\cite{voxposer}  & Jul. 2023 & Sim & \textcolor{teal}{\checkmark} & \textcolor{red}{$\times$}  & \textcolor{red}{$\times$}  & \textcolor{red}{$\times$}  \\
SayCan~\cite{saycan}      & Apr. 2022 & Real   & \textcolor{teal}{\checkmark} & \textcolor{red}{$\times$}  & \textcolor{teal}{\checkmark} & \textcolor{red}{$\times$}  \\
PALM-E~\cite{palme}       & Mar. 2023 & Sim, Real & \textcolor{teal}{\checkmark} & \textcolor{red}{$\times$}  & \textcolor{teal}{\checkmark} & \textcolor{teal}{\checkmark} \\
RT-2~\cite{rt2}           & Jul. 2023 & Real   & \textcolor{teal}{\checkmark} & \textcolor{red}{$\times$}  & \textcolor{teal}{\checkmark} & \textcolor{teal}{\checkmark} \\
SayPlan~\cite{rana2023sayplan}  & Jun. 2023 & Real   & \textcolor{red}{$\times$}  & \textcolor{red}{$\times$}  & \textcolor{teal}{\checkmark} & \textcolor{red}{$\times$}  \\
EmbodiedGPT~\cite{embodiedgpt}  & May 2023 & Sim    & \textcolor{teal}{\checkmark} & \textcolor{red}{$\times$}  & \textcolor{teal}{\checkmark}  & \textcolor{teal}{\checkmark}  \\
TaPA~\cite{tapa}          & Jul. 2023 & Sim   & \textcolor{red}{$\times$} & \textcolor{red}{$\times$}  & \textcolor{red}{$\times$}  & \textcolor{teal}{\checkmark}  \\
Voyager~\cite{wang2023voyager}      & May 2023 & Game   & \textcolor{red}{$\times$}  & \textcolor{teal}{\checkmark} & \textcolor{teal}{\checkmark} & \textcolor{red}{$\times$}  \\

Steve-Eye ~\cite{zheng2023steveeye} & Dec 2023 & Game   & \textcolor{teal}{\checkmark}  & \textcolor{red}{$\times$} & \textcolor{teal}{\checkmark} & \textcolor{teal}{\checkmark} \\

RoboScript~\cite{chen2024roboscript}      & Feb 2024  & Sim, Real   & \textcolor{red}{$\times$}  & \textcolor{teal}{\checkmark} & \textcolor{teal}{\checkmark} & \textcolor{red}{$\times$}  \\

% RoboCodeX~\cite{mu2024robocodex}      & Feb 2024  & Sim, Real   & \textcolor{teal}{\checkmark}  & \textcolor{teal}{\checkmark} & \textcolor{teal}{\checkmark} & \textcolor{teal}{\checkmark} \\
\midrule

Octopus     & - & Sim, Game & \textcolor{teal}{\checkmark} & \textcolor{teal}{\checkmark} & \textcolor{teal}{\checkmark} & \textcolor{teal}{\checkmark} \\ \bottomrule
\end{tabular}}
\vspace{-10pt}
\end{table}

More discussion on additional related works (Vision Language Model, and Feedback in Large Language Models) is included in the supplementary materials.

\section{The OctoVerse Environment}
\label{sec:octoverse}
In this section, we introduce three simulator environments designed to train and evaluate the Octopus model. For each environment, we will describe their overall information, the special design considerations that ensure the tasks are well-formulated and the callable functions are vision-dependent. 

\vspace{4pt}
\noindent\textbf{OctoGibson}\quad
% \href{https://docs.google.com/spreadsheets/d/1v-R5iekytQ5o_TAIathovpra-CxyCyN2WRKHDE2NvRw/edit?usp=sharing}{this google sheet}.}. 
We built the environment on the foundation of OmniGibson~\cite{li2023behavior}, an existing simulation framework that supports 1,000 daily activities across 50 scenes, featuring over 5,000 meticulously annotated objects. We incorporated 16 functions that the robot can execute, such as \texttt{moveBot()} and \texttt{easyGrasp()}.
Within this environment, we meticulously crafted 476 tasks\footnote{The full list of tasks and their categories are listed in the supplementary material.}, each with a well-defined initial state and a definitive termination state, allowing for a straightforward assessment of task completion.
Among these tasks, 367 are routine tasks—simple and direct actions like ``place a glass in a trash can,'' marked as \texttt{Follow}. Conversely, the remaining 109 are reasoning tasks that necessitate deeper comprehension. An example is ``buy a chocolate,'' where the agent needs to know to pick a chocolate bar from the shelf and then place it, along with money, on the checkout counter, denoted as \texttt{Reason} tasks.

To ensure our tasks are vision-aware, we deliberately constrain the usage of certain functions, such as \texttt{moveBot(object)}, which moves the agent in front of the given \texttt{object}. To avoid making the task too easy and vision-agnostic, we limit the given parameter to a predefined set of large objects, such as tables and cupboards, rather than small items like cups and glasses. In this case, if the robot wants to pick up a cup, it needs to recognize whether the cup is on the table or in the cupboard. A simple \texttt{moveBot(object)} call with an inappropriate parameter would cause a runtime error. The full list of the functions is in the Appendix.

\vspace{4pt}
\noindent\textbf{OctoMC}\quad
The OctoMC environment is built on Minecraft~\cite{Minecraft}, a popular platform for reinforcement learning and game agents. We integrated 6 functional actions and crafted 40 tasks\footnote{Detailed tasks are listed in the supplementary material.}, each designed to facilitate comprehensive observations and executions by the agent. These tasks are distributed across 10 different biomes, including indoor, outdoor, and underground settings, under varying weather conditions.

However, existing Minecraft environments often lack vision-formulated tasks and the necessary structure for vision-language programming. For example, in Voyager~\cite{wang2023voyager}, the \texttt{exploreUntil()} function allows the player to navigate directly to specific blocks without relying on vision information. This function works by randomly exploring the environment within a certain range until the desired object is found, at which point it returns a true value, enabling the agent to interact with the located object, such as gold blocks or trees. While effective, this approach is entirely automated and does not utilize visual information, making it unsuitable for our vision-based objectives.

To address this limitation, we crafted a vision-dependent exploration function, \texttt{teleport(yaw, distance)}, which operates within the robot's perceptual range. This function ensures that the agent's operations are vision-dependent and require active perception and navigation within the environment. For instance, to locate a tree block, the agent must actively navigate towards the direction of the forest, relying on visual cues to guide its exploration.

% Despite its popularity, Minecraft's voxel-based 3D environment and sandbox-focused gameplay pose challenges in transferring knowledge to real-world environments. The game's unique representations introduce limitations in terms of sim-to-real integration, task expansion, and state extension, affecting the applicability of Minecraft-based models in real-world scenarios. To address these concerns, we introduce OctoGTA, a more realistic open-world platform that better aligns with our vision-based objectives and real-world applicability.

\begin{figure}[t]
\centering
\includegraphics[width=\textwidth]{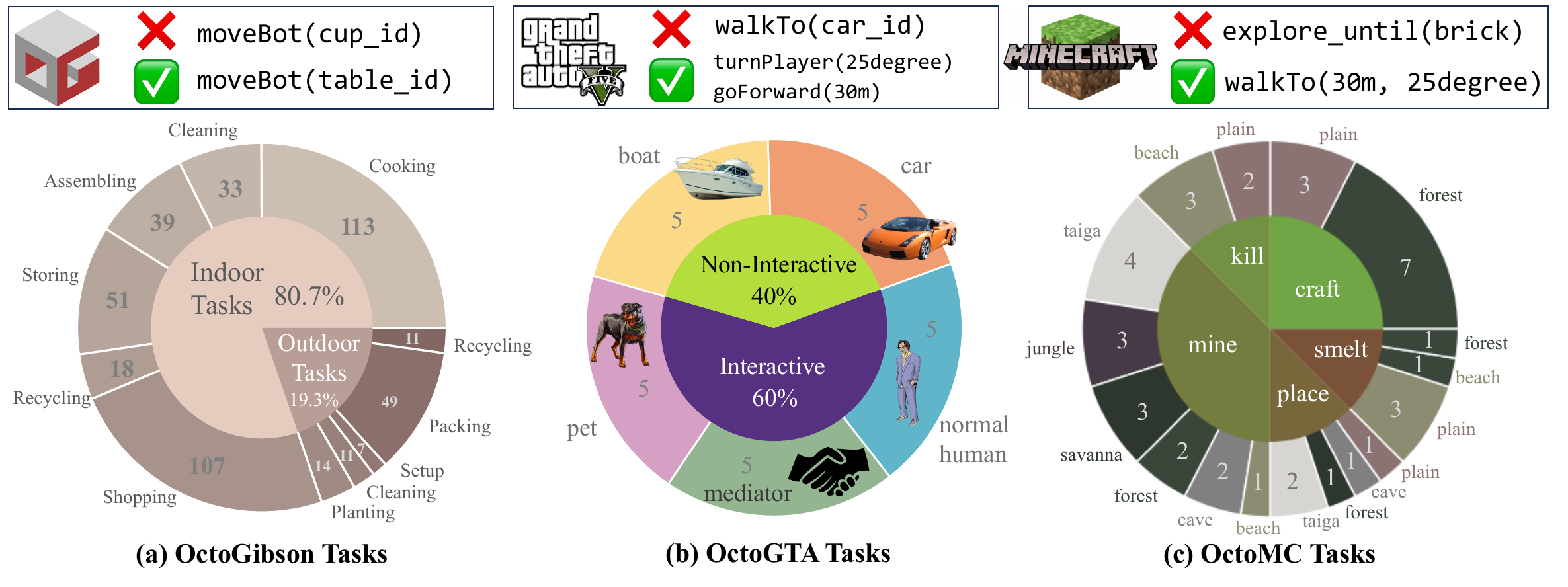}
\vspace{-20pt}
\caption{\textbf{The Statistics of the OctoVerse Environment with Function Designs.}}
\vspace{-10pt}
\label{fig:stat}
\end{figure}

\vspace{4pt}
\noindent\textbf{OctoGTA}\quad
The OctoGTA environment is built on GTA-V~\cite{gtav2014} with the help of the active GTA modding community. We integrated 19 functions and methodically crafted 25 tasks\footnote{We meticulously designed tasks to be friendly, ensuring they exclude any inappropriate or violent behaviors.}, such as ``help NPC drive their boat back to shore'' and ``mediate a fight between two NPCs,'' spanning across 5 groups (shown in Fig.~\ref{fig:stat} (b)). Each task is assigned to 5 different locations within the game world.

We have implemented a set of functions that enable the agent to interact with the game world in a visually-aware manner. Similar to the design in Minecraft, we get rid of functions like \texttt{walkTo(location)} that might trivialize the task of reaching a particular building or landmark. Instead, we provide functions such as \texttt{goForward(distance)} and \texttt{turnPlayer(degree)} (Fig.~\ref{fig:stat} (b)). For tasks like mediating a fight between two NPCs, the essential function \texttt{stopFight()} only works when the player is within 5 meters of the fighting NPCs. These design choices ensure that the agent's operations are vision-dependent and require active perception and navigation within the environment.

% \subsection{The OctoVerse Dataset}
% \textbf{The OctoGibson Dataset}\quad
% Following the operation in Section~\ref{sec:instructions} and \ref{sec:env_feedback}, we curated a training dataset within the OctoGibson environment. This training dataset encompasses 416 tasks, further divided into 3776 subtasks by GPT-4 exploration. For each subtask, beyond planning and executable code solutions, we capture 10 images representing the agent's perspective: 8 are egocentric images (spaced every 45 degrees), and 2 are bird's-eye view (BEV) images—one at a closer range and another at a greater distance. For evaluation purposes, we spare 60 tasks, of which 45 are routine tasks and 15 require reasoning. Additionally, 15 tasks are set in scenes not present in training.

% \textbf{The OctoGTA Dataset}\quad 
% The OctoGTA environment aims to validate the transferability of the Octopus model. Given the distinct code syntax from OctoGibson, we incorporated 9 tasks into the training set for few-shot adaptation, while reserving the remaining 11 tasks for evaluation. The training task encompasses 34 subtasks, each paired with 10 images and manually curated outputs.

%Each training task encompasses 5 subtasks, paired with 10 images and manually curated outputs.

\section{Octopus: The Embodied Vision-Language Programmer}
In this section, we present the procedure for training Octopus. Starting from collecting training data within the OctoVerse environment, Octopus builds upon a VLM architecture of Otter~\cite{li2023otter} and includes specialized RLEF modules to handle vision-language programming tasks. Fig.~\ref{fig:method} illustrates the entire Octopus training pipeline.

\subsection{Training Data Collection}
\label{sec:collection}
We use the automatic training data collection pipeline described here for OctoGibson and OctoMC, with the latter using customized prompts inspired by Voyager~\cite{wang2023voyager}. For OctoGTA, we rely on human labor to hand-craft the training dataset due to the difficulty of obtaining textual environment messages in the GTA environment. In the following parts, we use OctoGibson as the primary example to illustrate the data collection pipeline. Note that the primary task in organizing training data is to form a succinct pairing: ``vision input + current/historical states → next step plan + executable code''. 

\begin{figure}[t]
\centering
\includegraphics[width=\textwidth]{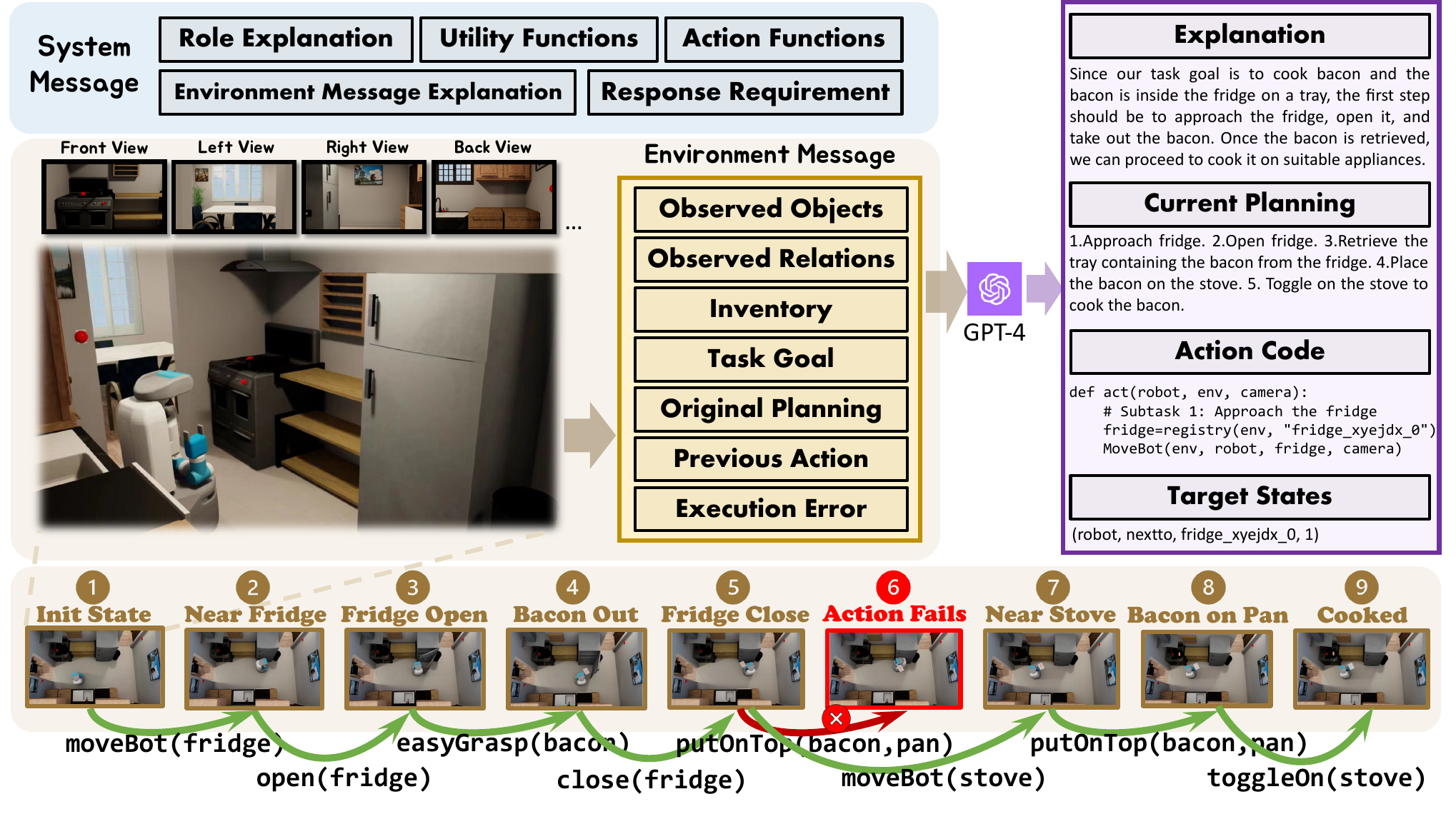}
\vspace{-25pt}
\caption{\textbf{Data Collection Example for ``Cook a Bacon'' Task.} GPT-4 perceives the environment through the \texttt{environmental message} and produces anticipated plans and code following the detailed \texttt{system message}. This code is subsequently executed in the simulator, directing the agent to the subsequent state. For each state, we gather the environmental message, wherein \texttt{observed objects} and \texttt{relations} are substituted by egocentric images to serve as the training input. The response from GPT-4 acts as the training output. Environmental feedback, specifically the determination of whether each target state is met, is documented for RLEF training.}
\label{fig:gpt4}
\vspace{-10pt}
\end{figure}

\begin{figure}[t]
    \centering
    \includegraphics[width=\linewidth]{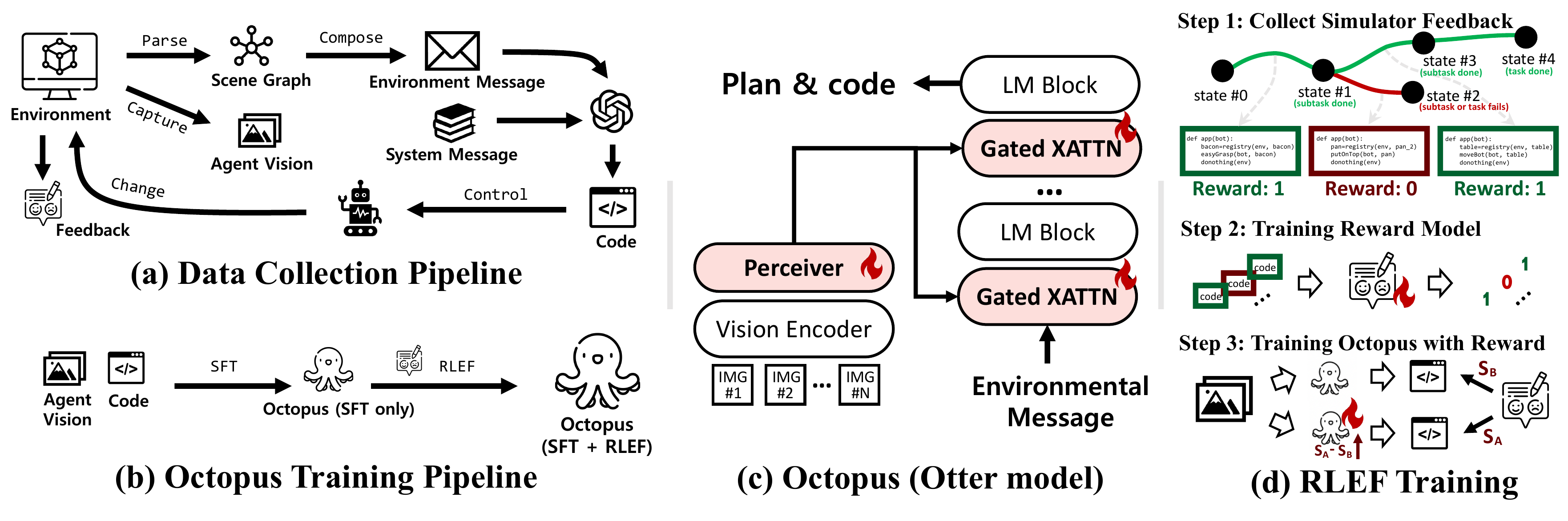}
    \vspace{-20pt}
    \caption{\textbf{How to train Octopus:} data collection and training pipeline.}
    \label{fig:method}
    \vspace{-15pt}
\end{figure}

% \noindent\textbf{Instructions From Exploration}\quad
% Initiating the training of the Octopus model involves ensuring its operational capability, particularly its ability to process vision input, interpret current and past states (such as objects the agent is holding), and produce structured plans and executable code.
% Thus, the primary task in organizing training data is to form a succinct pairing: ``vision input + current/historical states → next step plan + executable code''. However, collecting these pairs is far from simple; manually pairing them through human programmers would be both time-intensive and laborious. To circumvent this challenge, we harness the capabilities of GPT-4, not only to guide the agent's actions for task attempts but also to facilitate the automated data-gathering process.
\smallskip
\noindent\textbf{Environment Info Collection}\quad As shown in Fig.~\ref{fig:method} (a) and Fig.~\ref{fig:gpt4}, we format an \textbf{environment message} for each state, encompassing attributes like \texttt{Observed Objects}, \texttt{Observed Relations}, \texttt{Inventory}, and more. Specifically, the simulator can provide us with an exact scene graph at each state, shaping the content for the first two parts. The inventory info is also accessible. The task, e.g., ``cooking bacon'' in Fig.~\ref{fig:gpt4}, is represented by the \texttt{Task Goal}.

\smallskip
\noindent\textbf{Automation with GPT-4}\quad After preparing the environment message, we crafted a structured \textbf{system message} to ensure that the robot not only understands its input but also maintains a consistent output format. A detailed examination of this prompt can be found in the appendix. Experiments have shown that a well-articulated prompt enables GPT-4 to effectively generate executable code. It is important to note that the combined length of the system and environment messages can be extremely long, which may cause standard GPT-4 8K models to struggle with producing meaningful outputs. To address this issue, we employ the more robust GPT-4 32K model. As illustrated in Fig.~\ref{fig:gpt4}, when GPT-4 receives a consistent system and environment message, it generates comprehensive outputs that include current scenario analysis, planning, and actionable code, supporting the training process in Section~\ref{sec:sft}.

\smallskip
\noindent\textbf{Error Management}\quad 
Notably, GPT-4 collects training data under the main task of guiding the agent to complete tasks. However, GPT-4 is not infallible. Errors can manifest in multiple ways, ranging from syntax errors to physical challenges in the simulator. For instance, in Fig.~\ref{fig:gpt4}, between states \#5 and \#6, the action failed due to the long distance between the agent (holding bacon) and the pan. Such setbacks reset the task to its previous state. If a main task remains incomplete after 10 steps, it is deemed unsuccessful, and we terminate this task for budget concerns. However, all data pairs without syntax errors, regardless of the task's completion status, are valuable for refining instructions and improving the model's performance.

\smallskip
\noindent\textbf{Environmental Feedback}\quad
GPT-4's continual trial-and-error approach while guiding the agent toward task completion serves a dual purpose: collecting vision-output pairs and generating a rich set of feedback data.
The automatic annotation of this feedback focuses on two levels: step-level and task-level judgments. \textbf{Step-level judgment} assesses the alignment of post-execution states with their target states. For instance, in Fig.~\ref{fig:gpt4}, steps color-coded in green lead to positive feedback. One can visualize the action sequence for task completion as a tree, where each node indicates a step (subtask), encapsulating an action code. Accompanying each step is a binary value that denotes success or failure, giving preference to the successful branch over its counterpart. \textbf{Task-level judgment}, on the other hand, gauges the successful execution of the overall task. If the task is not completed as intended, every state within that task is labeled as negative, regardless of the status of the subtasks.
This collated feedback data serves as a foundation for our Reinforcement Learning with Environmental Feedback (RLEF) methodology, which we discuss in greater detail in Section~\ref{sec:rlef}.

\subsection{Model Architecture}
The Octopus architecture (shown in Fig.~\ref{fig:method} (c)) is heavily inspired by the Otter model~\cite{li2023otter}, integrates the \textbf{MPT-7B Language Decoder}\cite{MosaicML2023} and \textbf{CLIP VIT-L/14 Vision Encoder}\cite{radford2021learning}. Adopting design principles from Flamingo~\cite{alayrac2022flamingo}, Octopus employs the \textbf{Perceiver Resampler} and \textbf{Cross-Gated Attention modules} to enhance vision-language synergy. This architecture enables Octopus to excel in tasks requiring understanding of both visual and textual data. The Octopus is also compatible with other VLMs such as LLaVA~\cite{liu2023llava}.
% The Octopus architecture is heavily inspired by the foundation laid by the Otter model~\cite{li2023otter}. However, in our adaptation, specialized modifications have been made to tailor the architecture for the unique challenges of vision-language programming tasks found in OctoVerse. At the core of Octopus is the seamless integration of two critical components:
% \textbf{MPT-7B Language Decoder}~\cite{MosaicML2023} and \textbf{CLIP VIT-L/14 Vision Encoder}~\cite{radford2021learning}.

% To further enhance the synergy between the vision and language components, we have incorporated design principles from the Flamingo architecture~\cite{alayrac2022flamingo}. This is evident in our employment of the \textbf{Perceiver Resampler module} and the intricate weaving of \textbf{Cross-Gated Attention modules}. Initially, the Perceiver Resampler module ingests a sequence of image or video features to produce a fixed set of visual tokens. Subsequently, these tokens condition the language layers through Cross-Gated Attention modules, where the tokens act as keys and values while text from preceding layers serves as queries.

% Through this detailed architecture, the Octopus is primed to excel in tasks that demand a nuanced understanding of both visual and textual data.

\subsection{SFT: Supervised Finetuning with Instructions}
\label{sec:sft}
We train the Octopus model on our collected dataset from OctoVerse $\mathcal{D}_{\texttt{E}} = \{(\mathbf{X}_v, \mathbf{T}_i, \mathbf{T}_r)\}$ using token-level supervised fine-tuning (SFT)~\cite{ouyang2022training,touvron2023llama}. The Perceiver Resampler transforms images $\mathbf{X}_v$ into visual tokens that condition subsequent layers via Cross-Gated Attention modules. The training objective is next-token prediction, modeling the likelihood of a targeted response $\mathbf{T}_{r}$ as:
\begin{equation}
p(\mathbf{T}_{r} \mid \mathbf{T}_{i}, \mathbf{X}_{v}) = \prod^{L}_{l=1} p(t_{l} \mid \mathbf{X}_{v}, \mathbf{T}_{i}, \mathbf{T}_{r, < l}).
\end{equation}
Note that $\mathbf{T}_{i}$ denotes the instruction tokens and $\mathbf{T}_{r, < l}$ denotes the response tokens before the current predicted token $t_{l}$. During inference, tokens are converted into natural language via the language decoder's text tokenizer.

Visual observations $\textbf{X}_{v}=\{x^{0}_{F}, \ldots, x^{7}_{F}, x^{0}_{B}, x^{1}_{B}\}$ consist of 8 first-person view (FPV) images and two bird's-eye view (BEV) images for OctoGibson and OctoGTA. OctoMC only takes 4 FPV images. The FPV captures the agent's direct observations, while the BEV provides a holistic understanding of the environment. The eight FPV images are captured every 45 degrees, ensuring a complete 360-degree perspective.

\subsection{RLEF: Reinforcement Learning with Environmental Feedback}
\label{sec:rlef}
In OctoVerse, task progression can be visualized as a tree (Fig.~\ref{fig:method} (d)), where each node represents a sub-task with a binary value indicating success (1) or failure (0). If a node (or sub-task) has a value of 1, it is a step in the correct direction toward our end goal.

\noindent\textbf{Tree-based Task Representation} \quad
According to the environmental feedback part in Sec.~\ref{sec:collection}, environmental reward datasets $\mathcal{D}_{\texttt{R}} = {(\mathbf{X}^{*}_{v}, \mathbf{T}^{*}_{i}, \mathbf{T}^{i}_{r}, \mathbf{T}^{j}_{r}, c)}$ are organized, where $\mathbf{T}^{i}_{r}$ and $\mathbf{T}^{j}_{r}$ are responses sharing the same parent task $\mathbf{T}^{*}_{i}$, and $c$ indicates the preferred response leading to task completion. This ensures the reward mechanism favors the successfully executed branch. Note that even if a parental node does not have multiple responses, we can still assign feedback according to the rule in Sec.~\ref{sec:collection}.

\noindent\textbf{Reward Model Configuration} \quad 
A single-modal CodeLLaMA-7B model with an additional value head is fine-tuned on $\mathcal{D}_{\texttt{R}}$ as the reward model $r_{\phi}$. This text-based model assesses state transitions ($\mathbf{T}^{*}_{i} \rightarrow \mathbf{T}^{i,j}_{r}$) to determine high-reward transitions, assisting the agent in task execution and completion. The rationale for using CodeLLaMA as the reward model is that evaluating rewards can be purely dependent on the textual output. Furthermore, CodeLLaMA's strong programming skills make it well-suited for assessing the quality and effectiveness of the generated code in the context of task completion.

\noindent\textbf{Policy Model Development} \quad 
The supervised fine-tuned model serves as the initial policy model $\pi^{\texttt{INIT}}$ with fixed parameters. A duplicate model, $\pi^{\texttt{RL}}_{\theta}$, is initialized and trained using Proximal Policy Optimization (PPO)~\cite{schulman2017proximal} to maximize response rewards. The loss function is:
\begin{equation}
\mathcal{L}\left(\pi^{\texttt{RL}}_{\theta}\right) = -\mathbb{E}_{(\mathbf{X}^{*}_{v}, \mathbf{T}^{*}_{i}) \in \mathcal{D}_{\texttt{R}}, \mathbf{T}_{r} \sim \pi^{\texttt{RL}}} \left[ r_\phi(\mathbf{T}^{*}_{i}, \mathbf{T}_{r}) - \beta \cdot \mathbb{D}_\texttt{KL} \left( \pi^{\texttt{RL}}_{\theta} (\mathbf{X}^{*}_{v}, \mathbf{T}^{*}_{i}) \parallel \pi^{\texttt{INIT}}(\mathbf{X}^{*}_{v}, \mathbf{T}^{*}_{i}) \right) \right],
\end{equation}
where $\beta$ acts as a hyper-parameter to regulate the magnitude of the Kullback–Leibler (KL) penalty.

\section{Experiments}
\subsection{\diff{Main Results on OctoGibson}}
\noindent\textbf{Experimental Setup}\quad
We first set up the OctoGibson to evaluate the performance of Octopus and other related models. Specifically, we are utilizing the metrics of goal task completion score to check whether the task is completed in the simulator and the plan score from human evaluation. We have 60 evaluation tasks, with 45 from the seen environment, and 15 that are unseen during training. We also have 45 routine tasks and 15 require reasoning. Please note that models like Octopus might not always accurately identify specific object names as they appear in the simulator (e.g., ``water\_bottle\_189''). To address this, we implement a post-processing step for the generated code, substituting generic object references with their exact names from the simulator with simple string similarity matching. If multiple objects, we select the one closest to the agent. 

For Blind LLMs, we provide them with all the environment information in a textual format. Referring to Figure~\ref{fig:gpt4}, we hope the Blind LLMs could perform as GPT-4 but internalize the system message. For TAPA utilizes the open-vocabulary detection (OVD) technique~\cite{zhou2022detecting} to recognize objects within images and parse them into textual environmental messages, we still provide it with ground-truth environmental messages as an oracle setting.

\begin{table}[t]
\caption{\textbf{Main Results on OctoGibson.} We compare various models: standalone language models, adapted vision-language planners, and our Octopus models, across different evaluation settings. In cells displaying two values, the first represents the task completion rate across the target validation task sets, while the second assesses the conceptual accuracy of the model's planning as judged by human evaluators. GT denotes that the model input is directly parsed from the simulator, with information on objects (O) or relations (R). Octopus shows consistent advance in task completion.}
\label{tab:main_results}
\vspace{-10pt}
\centering
\resizebox{\textwidth}{!}{
\begin{tabular}{@{}lccccccc@{}}
\toprule
\multicolumn{1}{l}{\multirow{2}{*}{Model}} &
\multicolumn{1}{c}{\multirow{2}{*}{\begin{tabular}[c]{@{}c@{}}Vision\\ Model\end{tabular}}} &
\multicolumn{1}{c}{\multirow{2}{*}{\begin{tabular}[c]{@{}c@{}}Language\\ Model\end{tabular}}} &
\multicolumn{5}{c}{Entire Goal Task} \\ \cmidrule(l){4-8}
  \multicolumn{1}{c}{} &
  \multicolumn{1}{c}{} &
  \multicolumn{1}{c}{} &
  \multicolumn{1}{c}{Seen Env} &
  \multicolumn{1}{c}{Unseen Env} &
  \multicolumn{1}{c}{Follow} &
  \multicolumn{1}{c}{Reason} &
  \multicolumn{1}{c}{All} \\ \midrule
GPT-4 & - & -  &      0.42 / 0.69     &     0.46 / 0.67     &   0.49 / 0.78        &       0.27 / 0.40    &  0.43 / 0.68   \\
GPT-4V & - & - & 0.40 / 0.62  & 0.60 / 0.67  & 0.42 / 0.67  & 0.53 / 0.53 &  0.45 / 0.63   \\\midrule
LLaMA       & GT (O+R)  & LLaMA2-7B & 0.07 / 0.11 & 0.13 / 0.13 & 0.11 / 0.16 & 0.00 / 0.00 & 0.08 / 0.12 \\
CodeLLaMA  & GT (O+R)  & CodeLLaMA-7B  &   0.09 / 0.20        &    0.20 / 0.40       &     0.16 / 0.31      &    0.00 / 0.07       &   0.12 / 0.25  \\
TAPA (task-level)    & \sout{OVD} GT (O) & CodeLLaMA-7B &    0.09 / 0.36       &          0.13 / 0.33 &  0.11 / 0.36         &          0.06 / 0.33 &  0.10 / 0.35   \\
TAPA (step-level)        & \sout{OVD} GT (O) & CodeLLaMA-7B  &         \textbf{0.16} / \textbf{0.42}  &  0.13 / 0.27        &      \textbf{0.18} / 0.38     &  0.07 / 0.40         &   0.15 / 0.38  \\
% TAPA (step-level)        & OVD & CodeLLaMA-7B &     &        &         &           &        &    \\
\midrule
EmbodiedGPT & CLIP-ViT & MPT-7B  & 0.04 / 0.36          &      0.27 / \textbf{0.53}     &       0.13 / 0.38    &    0.00 / 0.40       &   0.10 / 0.40  \\
Octopus (SFT Only) & CLIP-ViT & MPT-7B  &      0.11 / 0.33     &     0.27 / 0.47     &   0.16 / 0.38        &       0.13 / 0.33    &  0.15 / 0.37   \\
Octopus (SFT + RLEF)    & CLIP-ViT & MPT-7B &    0.13 / 0.38       &       \textbf{0.33} / \textbf{0.53}    & \textbf{0.18} / \textbf{0.40}          &    \textbf{0.20} / \textbf{0.53}      &  \textbf{0.18} / \textbf{0.42}   \\
\bottomrule
\end{tabular}
}
\vspace{-20pt}
\end{table}

\smallskip
\noindent\textbf{CodeLLaMA Improves Coding but not Planning.} \quad
\diff{The first two rows in Table~\ref{tab:main_results} highlight the suboptimal task completion rate of the blind LLMs. Among them, CodeLLaMA boasts pre-training on a large programming dataset, resulting in a notable enhancement in code execution from our observation, with 92\% of the written code being successfully executed compared to LLaMA's 24\%. However, its prowess in planning remains limited. In contrast, the proposed Octopus MPT-7B model displays superior planning and task completion metrics while maintaining commendable coding abilities (72\% of the written code can be executed). We surmise that the coding requirements within the OctoGibson environment might not be exceedingly intricate, rendering an advanced programming language model, like CodeLLaMA, less crucial, albeit beneficial. For more insight, although not shown in the table, our efforts to replace the MPT model with CodeLLaMA encountered challenges of generating non-sense outputs, suggesting that more refined code, or image-code paired data might be necessary for a successful Octopus-CodeLLaMA integration.}

\smallskip
\noindent\textbf{Blind LLMs Struggle with Extended Input Content.} \quad
\diff{Our observations indicate that the step-level TAPA model, when supplied with a ground-truth object list, achieves a notable enhancement in planning. The primary distinction between it and the blind CodeLLaMA lies in the input length; the latter deals with protracted, pairwise relation content, complicating the language model's ability to extract crucial data from the environment message. This scenario highlights the inherent limitation of blind LLMs: relying on language alone to convey the entirety of environmental data can result in less informative input.}

\smallskip
\noindent\textbf{Octopus Demonstrates Superior Task Generalization.}\quad
\diff{Table~\ref{tab:main_results} underscores Octopus's strong performance, evidencing its consistent edge over standalone language models in task completion. Its adeptness in adapting to previously unencountered environments underlines the inherent advantages of vision-language models. A more detailed ablation analysis is provided later.}

\begin{figure}[t]
    \centering
    \includegraphics[width=\textwidth]{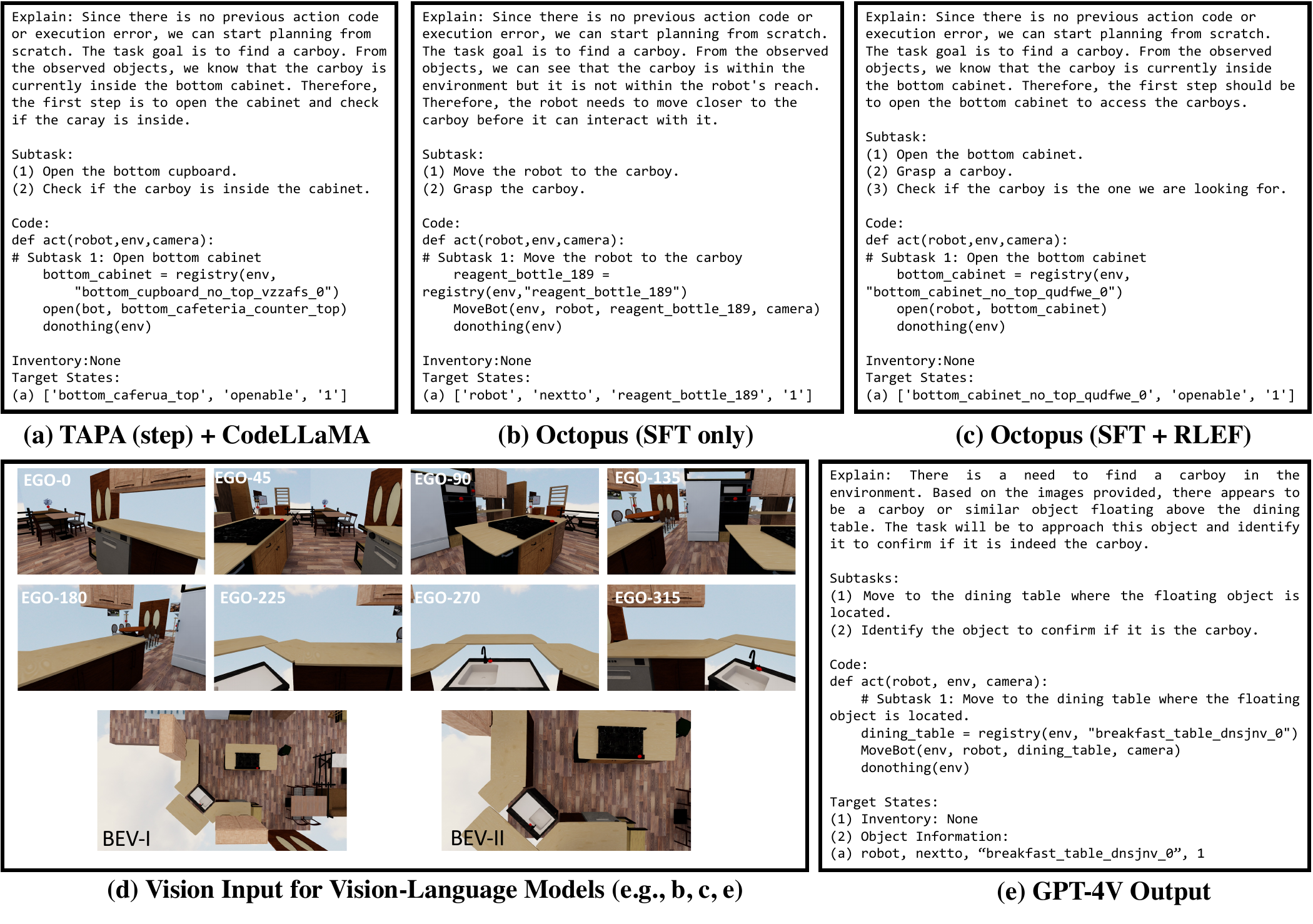}
    \vspace{-25pt}
    \caption{Qualitative Results on the task of \textit{find a carboy} in OctoGibson environment. We show that the models shown can write executable code, but the proposed Octopus has stronger planning ability, especially after RLEF. We also explore the performance of GPT-4V on the specific task.}
    \label{fig:code}
    \vspace{-20pt}
\end{figure}

\smallskip
\noindent\textbf{RLEF Enhances Octopus's Planning Strategy.}\quad
\diff{Table~\ref{tab:main_results} shows Octopus's strong reasoning capabilities after the RLEF finetuning. An example can be observed in Fig.~\ref{fig:code}(b-c), where, after refinement via RLEF, Octopus astutely navigates to the cabinet housing the carboy instead of attempting a direct yet distant capture. Quantitatively, Octopus exhibits enhanced adaptability to previously unseen reasoning tasks, reinforcing its prowess in logical task resolution. When juxtaposed with other strategies, such as the embodied queries employed by EmbodiedGPT, RLEF emerges as the more efficacious approach.}

\subsection{Ablation Study}

\begin{figure}[t]
    \centering
    \includegraphics[width=\textwidth]{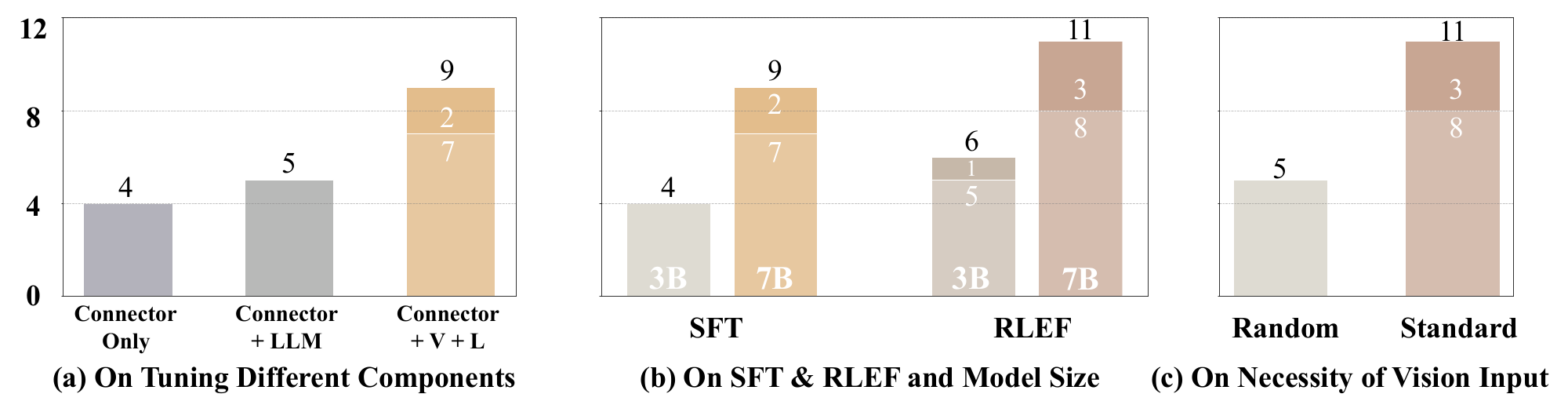}
    \vspace{-25pt}
    \caption{Ablation Study on model components, model size, and vision input. For bars with different colors, the upper bar denotes the number of successful reasoning tasks, and the lower is routine tasks.}
    \vspace{-20pt}
    \label{fig:ablation}
\end{figure}

\noindent\textbf{Tunning Different Components} \quad 
\diff{Fig.~\ref{fig:ablation} (a) demonstrates that solely adjusting the connector (marked ``fire'' in Fig.~\ref{fig:method} (a)) leads to success for merely 4 out of 60 tasks. Conversely, finetuning both the connector and language decoder nudges the success rate slightly higher, with 5 tasks being accomplished.}

\smallskip

\noindent\textbf{7B \textit{v.s.} 3B Model Size} \quad 
\diff{We embarked on experiments centered on model size to discern the influence of the total parameter count on the efficacy of vision-language models. As illustrated in Fig.~\ref{fig:ablation} (b), downsizing the model manifests in a noticeable performance drop. The congruency of results across both the SFT and RLEF models underscores the importance of an apt model size when sculpting vision-language models.}

\smallskip
\noindent\textbf{Significance of Visual Inputs in Task Performance}\quad
\diff{In our standard configuration, the vision component processes a sequence of image inputs, consisting of eight circularly captured first-person view (FPV) images, complemented by two bird's-eye view (BEV) images. With the intent to investigate the impact of visual inputs on task performance, we initiated an ablation study. In a modified setup, the sequence of these visual inputs was deliberately randomized, aiming to attenuate the strength of the visual signals. As illustrated in Fig.~\ref{fig:ablation} (c), this intentional disruption in visual input consistency led to a pronounced decline in task performance. This result highlights the crucial role that clear and structured visual inputs play in the Octopus model, emphasizing that it significantly leverages visual cues for effective planning and task execution.}

% \subsection{Performance of GPT-4 and GPT-4V}
% \noindent\textbf{Performance of GPT-4}\quad 
% \diff{The input provided to GPT-4 was consistent with the input during our data collection phase, which was purely textual. Under such conditions, out of a total of 60 test tasks, GPT-4 achieved a commendable success rate in 31 tasks. This result suggests that current models still possess considerable room for advancement. The fact that even GPT-4 doesn't perform optimally indicates a vast scope for improvements within the domain.}

% \noindent\textbf{Performance of GPT-4V}\quad 
% \diff{Though we couldn't extensively test GPT-4V due to API limitations, our sample case indicates its ability to generate code on par with Octopus when provided with image-based environment messages. However, while Octopus, having been trained in the present environment, adeptly performs tasks like ``open the cabinet'', GPT-4V's actions, shown in Fig.~\ref{fig:code} (e), although seemingly accurate, fall short in specific tasks such as locating the target object - the carboy. Given GPT-4V's zero-shot learning approach and its unfamiliarity with our environment, alongside potential simulator discrepancies, its results remain commendable.}

\subsection{Results on Minecraft and GTA Tasks}

\begin{table}
\caption{\textbf{Main Results for GTA and Minecraft Tasks.} Despite limited training data, the Octopus still shows its good ability in task completion. In cells displaying two values, the first represents the task completion rate across the target validation task sets, while the second assesses the conceptual accuracy of the model's planning as judged by human evaluators.}
\label{tab:game}
\label{tab:game}
\centering
\vspace{-8pt}
\resizebox{\linewidth}{!}{
\begin{tabular}{@{}lcccccccc@{}}
\toprule
\multicolumn{1}{l}{\multirow{2}{*}{Model}} &
\multicolumn{1}{c}{\multirow{2}{*}{\begin{tabular}[c]{@{}c@{}}Vision\\ Model\end{tabular}}} &
\multicolumn{1}{c}{\multirow{2}{*}{\begin{tabular}[c]{@{}c@{}}Language\\ Model\end{tabular}}} &
\multicolumn{3}{c}{OctoMC}&\multicolumn{3}{c}{OctoGTA}\\ 
\cmidrule(l){4-9}
  \multicolumn{1}{c}{} &
  \multicolumn{1}{c}{} &
  \multicolumn{1}{c}{} &
  \multicolumn{1}{c}{Seen Task} &
  \multicolumn{1}{c}{Unseen Task} &
  \multicolumn{1}{c}{All} &
  \multicolumn{1}{c}{Seen Task} &
  \multicolumn{1}{c}{Unseen Task} &
  \multicolumn{1}{c}{All} 
  \\ \midrule
  GPT-4 & - & - & 0.70 / 0.85 & 0.60 / 0.80 & 0.65 / 0.83 & 0.55 / 0.80 & 0.50 / 0.60 & 0.54 / 0.76\\
  % \midrule
  % LLaVA (zero-shot)  & - & - & 0.25 / 0.45 & 0.20 / 0.40 & 0.23 / 0.43 & 0.13 / 0.30 & 0.10 / 0.30 & 0.12 / 0.30\\
  GPT-4V & - & - & 0.75 / 0.85 & 0.70 / 0.85 & 0.73 / 0.85 & 0.58 / 0.85 & 0.50 / 0.70 & 0.56 / 0.82\\
  \midrule
  EmbodiedGPT & CLIP-ViT & MPT-7B & 0.30 / 0.55 & 0.20 / 0.60 & 0.25 / 0.58 & 0.15 / 0.38 & 0.20 / 0.60 & 0.16 / 0.42\\
  Octopus (SFT Only) & CLIP-ViT & MPT-7B      &      0.30 / 0.60     &     0.20 / 0.70     &   0.25 / 0.65        &       0.18 / 0.48    &  0.20 / 0.60  & 0.18 / 0.50 \\
  Octopus (SFT + RLEF)    & CLIP-ViT & MPT-7B &    0.40 / 0.60      &       0.20 / 0.70    & 0.30 / 0.65  &  0.18 / 0.53      &  0.30 / 0.70  & 0.20 / 0.56\\
  % Octopus (LLaVA, SFT Only) & CLIP-ViT & Vicuna-7B  &  0.35 / 0.50 & 0.30 / 0.65& 0.33 / 0.58 & 0.20 / 0.53 & 0.20 / 0.70& 0.20 / 0.56\\
\bottomrule
\end{tabular}
}
\vspace{-25pt}
\end{table}

\smallskip

\noindent\textbf{Results on OctoMC}\qquad
According to the OctoMC part in Sec.~\ref{sec:octoverse}, we designed 40 tasks, each task is operated on 2 locations so a total of 80 tasks. 
We set aside 10 tasks as unseen tasks and 10 tasks as seen tasks, getting 60 training tasks and 30 testing tasks in OctoMC.
Similar to OctoGibson, the training data for OctoMC is collected using GPT-4. The agent, guided by GPT-4, explores the Minecraft environment and generates action plans and corresponding code based on the provided system messages, environmental cues, and objectives.

Table~\ref{tab:game} shows that the SFT model trained on OctoMC can complete most tasks in both seen and unseen scenarios, demonstrating better performance compared to OctoGTA. However, upon analyzing the failure cases, we find that the model sometimes struggles with tasks requiring precise spatial reasoning. For instance, when tasked with killing a pig, the agent may have difficulty finding the creature with the exact angle and distance. While it shows that the agent relies on visual information to navigate and interact with the environment, it also means that even with correct planning, imprecise actions can lead to failure.

\noindent\textbf{Results on OctoGTA}\qquad
According to the OctoGTA part in Sec.~\ref{sec:octoverse}, we designed 25 tasks. We set aside 5 tasks in the boat-related group (e.g., boat retrieval and shore return) as unseen tasks for testing only, using 2 different locations. We replicate the remaining 20 tasks for both training (8 different locations) and testing (2 locations). As a result, we have 160 training tasks and 50 testing tasks in OctoGTA. Unlike the training procedure for OctoGibson and OctoMC, the training data for OctoGTA is entirely created by the authors, as it is challenging to gather textual environmental messages in the GTA environment.

Table~\ref{tab:game} shows that, despite having only 160 training tasks, the SFT model can complete some tasks in both seen and unseen scenarios, and RLEF also outperforms. However, upon careful examination of the failure cases, we find that the model struggles with tasks that are not straightforward. For instance, when a wall separates the player from the car, the player still finds it difficult to decide to climb the wall, even if similar cases exist in the training data. Similar to OctoMC, as illustrated in Sec.~\ref{sec:octoverse}, when approaching certain locations, the code involves functions like \texttt{turnPlayer()} and \texttt{goForward()} rather than a simple \texttt{walkTo(location)}. Consequently, even with correct planning, imprecise actual actions can still lead to task failure.

\section{Conclusion}
This paper introduces Octopus, an embodied vision-language programmer designed to bridge the gap between high-level planning and real-world manipulation with programming. By open-sourcing our OctoVerse environments, dataset, and Octopus architecture, we aim to foster collaboration and innovation within the research community, paving the way for future developments in embodied vision-language programming.

\section*{Acknowledgments and Disclosure of Funding}
This research/project is supported by the National Research Foundation, Singapore under its AI Singapore Programme (AISG Award No: AISG2-PhD-2022-01-029). Besides, this study is supported by the Ministry of Education, Singapore, under its MOE AcRF Tier 2 (MOET2EP20221- 0012), NTU NAP, and under the RIE2020 Industry Alignment Fund – Industry Collaboration Projects (IAF-ICP) Funding Initiative, as well as cash and in-kind contribution from the industry partner(s).

\bibliographystyle{splncs04}
\bibliography{egbib}

\newpage
\appendix
\input{appendix}

% \bibliography{egbib}

\end{document}

%% file: appendix.tex
\appendix
\setcounter{table}{0}
\renewcommand{\thetable}{A\arabic{table}}
\setcounter{figure}{0}
\renewcommand{\thefigure}{A\arabic{figure}}

\section{OctoGibson}
\subsection{The difference between OctoGibson and OmniGibson}
OctoGibson builds upon the foundation of OmniGibson, a simulation framework that supports a wide range of daily activities across diverse scenes with numerous annotated objects. However, OctoGibson extends OmniGibson in several crucial ways to support embodied vision-language programming. 

\noindent\textbf{Add Controllable State for Objects}\quad Each object's operable properties are described by 8 unary states, such as \textbf{openable} and \textbf{heatable}, 

\noindent\textbf{Add Relation Parser}\quad The OctoGibson adds 12 binary relations, such as \textbf{next to} and \textbf{on top}, to illustrate its spatial relationships with other objects. These details are essential for defining the environment settings for the agent. 

\noindent\textbf{Add Tasks}\quad
OctoGibson introduces a set of 476 meticulously crafted tasks, each with well-defined initial and goal states, enabling clear evaluation of task completion. These tasks are categorized into routine tasks that involve simple, direct actions, and more complex reasoning tasks that require multi-step planning. 

\noindent\textbf{Add Function Calls}\quad
OctoGibson incorporates 16 carefully designed functions that the agent can execute, such as moveBot() and easyGrasp(), to interact with the environment in a more structured manner. 

\noindent\textbf{Add Visual-Dependent Function Calls}\quad
to ensure that the agent's actions are grounded in visual perception, OctoGibson imposes certain constraints on the function parameters, such as limiting moveBot() to only accept large, fixed objects as arguments. This encourages the agent to reason about the scene and plan accordingly, rather than relying on hard-coded knowledge. 

Together, these enhancements make OctoGibson a more suitable platform for studying embodied vision-language programming compared to the base OmniGibson environment.

\subsection{OctoGibson Dataset}
% % data table @cho
The OctoGibson training dataset comprises 476 tasks, further subdivided into 3,776 instructional subtasks. Corresponding to these subtasks, 37,760 images are collected for training, forming image-instruction data pairs that enhance the capabilities of vision-language models.

\begin{table}[h!]
\caption{The Statistical Overview of the OctoGibson Dataset.}
\vspace{2pt}
\centering
\label{tab:OctoGibson}
\setlength\tabcolsep{12pt}
\renewcommand{\arraystretch}{1.4}
\resizebox{1.0\textwidth}{!}{
\begin{tabular}{@{\hskip 0.05in}l@{\hskip 0.2in}ccl@{\hskip 0.05in}}
\toprule
Dataset &Type & Number & Comments \\
\midrule
\multirow{6}{*}{\textbf{OctoGibson}}
& Objects &78,138& Objects are divided into 428 categories. (E.g. \textbf{pork}, \textbf{scanner}, \textbf{sofa}, \textbf{sweater}) \\
& States &8& States represent the operable properties of an object. (E.g. \textbf{openable}, \textbf{heatable})\\
& Relations &12& Relations describe the spatial relations between two objects. (E.g. \textbf{nextto}, \textbf{ontop}) \\
& Images & 37,760 &The images are captured in an 80\% egocentric and 20\% bird's-eye view perspective  \\
& Layout &16 &Layout provides task environments: \textbf{Interior Scene}, \textbf{Outdoor Scene}, and \textbf{Public Scene}.\\
& Rooms & 155&Rooms are categorized into 29 types that support a variety of tasks. (E.g. \textbf{garage}, \textbf{child's room}, and \textbf{dining room})\\
\bottomrule
\end{tabular}
}
\end{table}

\subsection{How We Collect Training Data}
Following Fig. 3 in the main paper, we use GPT-4 to automatically collect responses using the system message and environment message shown below.

\lstset{ 
    breaklines=true,
    breakatwhitespace=true,
    postbreak=\mbox{\textcolor{red}{$\hookrightarrow$}\space},
    columns=fullflexible,
    keywordstyle=\color{blue!70},
    commentstyle=\color{red!50!green!50!blue!50},
    basicstyle=\ttfamily,
    frame=single,
} 
\flushleft{
\noindent\textbf{System Message}
\begin{lstlisting}
You are a vision language assistant agent with high intelligence. 

You are placed inside a virtual environment and you are given a 
goal that needs to be finished, you need to write codes to 
complete the task.

You can solve any complex tasks by decomposing them into subtasks 
and tackling them step by step, but you should only provide 
the action code for solving the very next subtask, because the 
action code needs time to be compiled and executed in the simulator 
to check whether they can be operated successfully.

Here are some useful programs that you may need to use to complete the tasks.

You need to use the utility functions to complete the tasks.

Utility Functions:
donothing(env): wait for the system to capture.
registry(env, obj_name): each time you want to use an object in the environment, call this function first. obj(str): the object in the environment. e.g. apple_1234 = registry(env,"apple_1234"), then you can use apple_1234 to represent "apple_1234" in the environment. For each object, you can only register it once, don't register an object multiple times. By default, the variable name should be the same as the string.

The Action List contains multiple defined functions, you could execute your actions by calling these functions.
I will first give you the name of the function as well as its input, then I will give you an explanation of what it can do, e.g. function_name(inputs): capability of the function.

Action List:
EasyGrasp(robot, obj): The robot will grasp the object.
MoveBot(env, robot, obj, camera): Move the robot in the env to the front of obj. Note that the robot can only move to a position in front of large objects (e.g.,  tables, ovens, etc.) that are placed directly on the ground. The robot cannot directly move to small objects (e.g., apples, plates, etc.). The camera should always be set to camera.
put_ontop(robot, obj1, obj2): Put the obj1 within the robot's hands onto obj2
put_inside(robot, obj1, obj2): Put the obj1 within the robot's hands inside obj2
cook(robot,obj): cook the given object.
burn(robot,obj): burn the given object.
freeze(robot,obj): freeze the given object.
heat(robot,obj): heat the given object.
open(robot,obj): open the given object.
close(robot,obj): close the given object.
fold(robot,obj): fold the given object.
unfold(robot,obj): unfold the given object.
toggle_on(robot,obj): toggle on the given object.
toggle_off(robot,obj): toggle off the given object.

At each round of conversation, I will give you
Observed Objects: ...
Observed Relations: ...
Inventory: ...
Task Goal: ...
Original Subtasks: ...
Previous Action Code: ...
Execution Error: ...

I will give you the following information for you to make a one-step action decision toward the final goal.
(1) Observed Objects: contains object names, its editable states with the corresponding value of the states and distance measuring the centroid of Agent towards the object. It denotes with (object, [(state1, value1), (state2, value2)], distance).e.g. (fridge, [('openable', 1)], 1.8) means the object fridge can be opened, and it is currently openedand and the distance is a float value measured in meters.
(2) Observed Relations: a scene relation graph triplet denotes with (object, relation, object), e.g. (apple, ontop, desk). You are termed with Agent in this context.
(3) You should pay attention to the relation graph which is essential for you to understand the status of the environment.
(3) The observation may not include all the information about the objects you need to interact with, the objects may be hidden inside other objects, so feel free to explore the reasonable place they might appear.
(4) The Inventory contains a stack-like structure, you could put things inside. But remember first in last out.  It contains all the things the robot has in its hand. If nothing is in Inventory, denoted with None.
(5) The Task Goal contains instructions and the Agent finished state for the entire goal.
(6) Original Subtasks: The sub-tasks that is planned in the conversation. Note that the original plans could be problematic and unable to solve the problem, so you might need to make revision and set up a new plan if necessary.
(7) Previous Actions: The action code for solving the previous subtasks would be provided so that you can understand what was going on and extend the code with the action code for solving the next subtask. Pay attention to the number used in camera functions in previous code, make sure the number is continuous.
(8) Execution Error: The execution error for last round will be provided to help you in this round.

You should then respond to me with
Explain (if applicable): Are there any steps missing in your plan? Why does the code not complete the task? What does the chat log and execution error imply?

Subtasks: How to complete the Task Goal step by step by calling given action functions. You should plan a list of subtasks to complete your ultimate goal. You need to make the planning consistent to your previous round unless those need to change. You should pay attention to the Inventory since it tells what you have. The task completeness check is also based on your final inventory. Pay attention that you can only interact with the objects within two meters of you, so you need to be close enough to interact with the objects.

Code:
(1) Remember you can only interact with the objects within two meters of you.
(2) Only use functions given in Utility Functions, Action List. Write a function taking the 'robot', 'env' and 'camera' as the only three arguments.
(3) Reuse the above useful programs as much as possible.
(4) Your function will be reused for building more complex functions. Therefore, you should make it generic and reusable. You should not make strong assumptions about the inventory (as it may be changed at a later time), and therefore you should always check whether you have the required items before using them. If not, you should first collect the required items and reuse the above useful programs.
(5) The function name should always be 'act', but you need to explain what task it completes.
(6) Each time you take an action in the provided action list, after you take the action, you have to use the function 'donothing' before you take another action in the action list. So the block should look like "One action in the action list + donothing". Remember one action in your plan may contain multiple actions in the action list, you have to use the block for each action in the action list.
(7) Registry every object you might need to use first.
(8) You should only output the action code to finish your very next subtask. Remember not to generate the entire action code unless it is the final step.
(9) You can have more than one things in Inventory.

Also please notice that registration should not be considered as one subtask. Make sure that your subtask planning should start with real actions like "open the door" while keeping the object registry as the default action.

Target States: A state to check the completeness of the subtask. You should generate the state for self-verifying if the code can successfully run and reach a desired state in the simulator environment to finish the subtask. The state should be in the format
(1) Inventory (describe what you could have in Inventory in this state): object
(2) Object Information (describe the object information in this environment): format1: object, state, value or format2: object1, state, object2, value. The value can only be 0 or 1, representing False or True of the state. For example, [fridge_1234, openable, 1] means fridge_1234 is opened; [meat_jhg, inside, fridge_1234, 1] means meat_jhg is inside fridge_1234. For format1, you can only choose the state from: ['cookable', 'burnable', 'freezable', 'heatable', 'openable', 'togglable', 'foldable', 'unfoldable']. For format2, you can choose the state from: ['inside', 'nextto', 'ontop', 'under', 'touching', 'covered', 'contains', 'saturated', 'filled', 'attached', 'overlaid', 'draped']. If the object is the robot, denote it with 'robot'.
(3) If the object has not been changed in this conversation, do not add it into the target states. 
(4) You don't need to write any annotations for target states.
(5) Remember to make sure the states you use is in the provided state list for format1 and format2.
(5) You can only use the objects provided in the Object Information part, you cannot use the name you registered in the code.
(6) The object information of target states should be the last part of your response, no more explanations are needed.

## Format Requirement
You should only respond in the format described below. Please strictly pay attention to the format of the bullet points, especially the brackets for the number (e.g., "(1), (2), and (3)").
{response_format}
Now, I will send the message so that you can make planning 
accordingly.

Explain: 
...
Subtasks:
(1) ...
(2) ...
(3) ...
// Please provide me with ALL previous subtasks (E.g if subtask1 & 2 are successfully acted and make mistakes in subtask3, please return me with subtask1 & 2 and new plan of subtask3)
...
Code:
```python
// import neccessary modules first
// helper functions (only if needed, try to avoid them)
...
// main function after the helper functions
def act(robot,env,camera) {
  // comment should be clear and correspond to subtasks above (e.g., Subtask 1: xxxx)
  //only generate one subtask in each act function
}
```
Target States:
(1) Inventory: ... //robot can grasp more than one thing at the same time
(2) Object Information:
    (a) ... // e.g. (a) table, nextto, fridge, 1
    (b) ...
    (c) ...
\end{lstlisting}

\noindent\textbf{Environmental Message}
\begin{lstlisting}
    Observed Objects: (bacon_150, (['heatable', 0], ['freezable', 0], ['burnable', 0], ['cookable', 0]), 1.89)(fridge_xyejdx_0, (['openable', 0], ['heatable', 0], ['freezable', 0]), 2.12)(stove_rgpphy_0, (['togglable', 0], ['heatable', 0], ['freezable', 0]), 1.59)(tray_156, (['heatable', 0], ['freezable', 0]), 1.85)(griddle_157, (['togglable', 0], ['heatable', 0], ['freezable', 0]), 1.68)
    Observed Relations: [('bacon_150', 'inside', 'tray_156'), ('tray_156', 'inside', 'fridge_xyejdx_0'), ('bacon_150', 'inside', 'fridge_xyejdx_0')]
    Inventory: None
    Task Goal: cook_bacon
    Original Subtasks: None
    Previous Action Code: No code
    Execution error: No error
    Now, please output Explain, Subtasks (revise if necessary), Code that completing the next subtask, and Target States, according to the instruction above. Remember you can only use the functions provided above and pay attention to the response format.
\end{lstlisting}

\noindent\textbf{GPT-4 Response Example}
Here we also show the example response that is collected from GPT-4.
\begin{lstlisting}
Explain:
Since our task goal is to cook bacon and the bacon is inside the fridge on a tray, the first step should be to approach the fridge, open it, and take out the bacon. Once the bacon is retrieved, we can proceed to cook it on a suitable appliance like the griddle or stove.

Subtasks:
(1) Approach the fridge.
(2) Open the fridge.
(3) Retrieve the tray containing the bacon from the fridge.
(4) Approach the or stove.
(5) Place the bacon on the stove.
(6) Toggle on the stove to cook the bacon.

Code:
def act(robot, env, camera):
    # Subtask 1: Approach the fridge
    fridge_xyejdx_0 = registry(env, "fridge_xyejdx_0")
    MoveBot(env, robot, fridge_xyejdx_0, camera)
    donothing(env)
    
Target States:
(1) Inventory: None
(2) Object Information:
(a) robot, nextto, fridge_xyejdx_0, 1
\end{lstlisting}

\subsection{Performance of GPT-4 and GPT-4V}
\noindent\textbf{Performance of GPT-4}\quad 
\diff{The input provided to GPT-4 was consistent with the input during our data collection phase, which was purely textual. Under such conditions, out of a total of 60 test tasks, GPT-4 achieved a commendable success rate in 31 tasks. This result suggests that current models still possess considerable room for advancement. The fact that even GPT-4 doesn't perform optimally indicates a vast scope for improvements within the domain.}

\noindent\textbf{Performance of GPT-4V}\quad 
\diff{Though we couldn't extensively test GPT-4V due to API limitations, our sample case indicates its ability to generate code on par with Octopus when provided with image-based environment messages. However, while Octopus, having been trained in the present environment, adeptly performs tasks like ``open the cabinet'', GPT-4V's actions, shown in Fig.5 (e), although seemingly accurate, fall short in specific tasks such as locating the target object - the carboy. Given GPT-4V's zero-shot learning approach and its unfamiliarity with our environment, alongside potential simulator discrepancies, its results remain commendable.}

\section{OctoMC}
\begin{figure}[t]
    \centering
    \includegraphics[width=\linewidth]{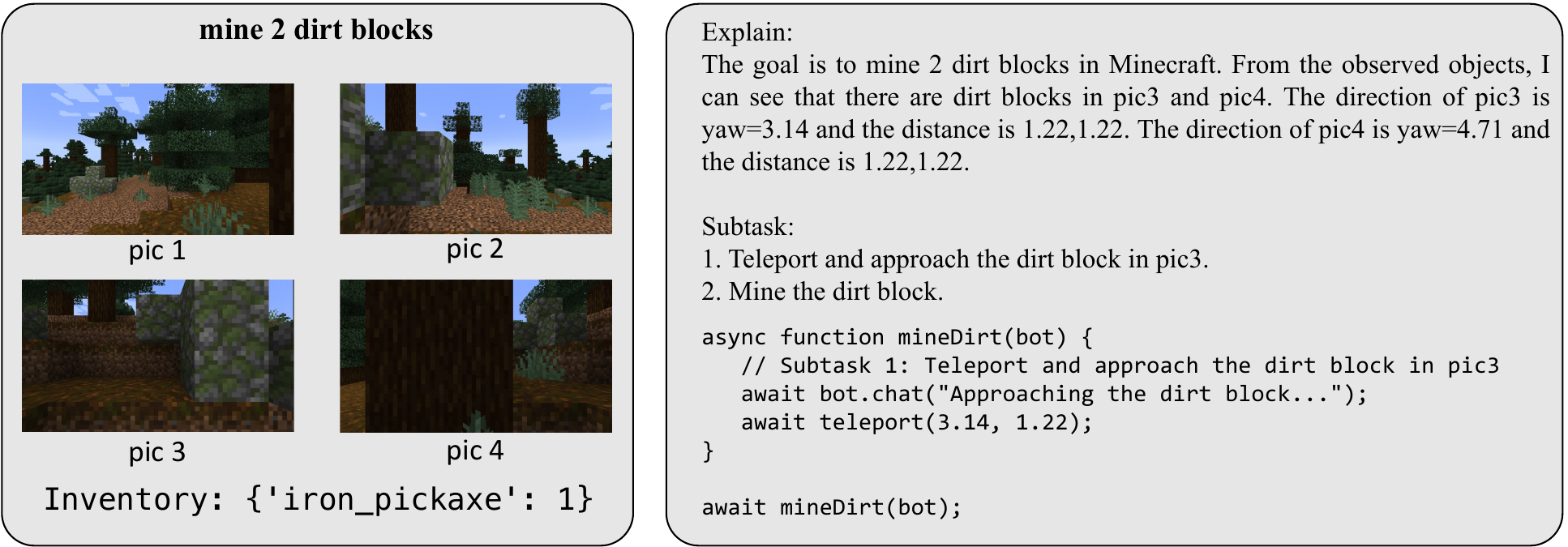}
    \caption{Example of Minecraft Code. Input on the left, output on the right.}
    \label{fig:mc_code}
\end{figure}
\subsubsection{Background}\quad
In recent years, Minecraft's open-ended environment has garnered significant attention in the field of reinforcement learning and game agents research. The advent of Large Language Models (LLMs) and Large Multimodal Language Models (LMMs) has introduced a new dimension to this domain, enabling agents to generate executable plans or policies across a broad spectrum of skills and tasks within open-ended worlds like Minecraft. However, existing Minecraft environments often lack vision-formulated tasks and the necessary structure for vision-language programming.

To address this gap, we introduce our secondary environment, OctoMC, built upon the foundation of Minecraft~\cite{Minecraft}. OctoMC is designed to provide a set of function calls and tasks that leverage constructed vision information across different weather conditions and biomes. We utilized the high-level JavaScript API provided by Mineflayer\footnote{https://github.com/PrismarineJS/mineflayer} to extract visual information from the Minecraft world. The function \texttt{bot.canSeeBlock(block)} performs raycasting around the bot to determine the visibility of specific blocks, while \texttt{bot.blockAt(block)} identifies surrounding blocks through iterative searching.
A comprehensive study of the Minecraft-based work is listed in Table~\ref{minecraft-agent}, showing that OctoMC is tailored for VLM programming and vision-aware function calls.

\subsection{Highlighting Vision-based Function Call}
Building upon these capabilities, we crafted a vision-dependent exploration function, \texttt{teleport(yaw, distance)}, which operates within the robot's perceptual range. This function identifies the target block and computes the distance from the bot entity to the target, utilizing lidar properties alongside the occupancy and optical characteristics of the Minecraft ego-view camera to enhance vision-based navigation and interaction within the game environment.
% \subsubsection{Minecraft Benchmark}
% In recent years, the 3D sandbox Minecraft has received considerable attention owing to its remarkably flexible game mechanics to serve as a prominent open-world benchmark (e.g., MineRL (Guss
% et al., 2019) and Minedojo (Fan et al., 2022)).
\begin{table} 
\captionsetup{font=scriptsize}
\begin{center} 
\scriptsize

\caption{\textbf{Related Models and Methods for Minecraft Agents}\quad This summary describes the methods used by Minecraft agents, focusing on how they combine Language/Vision, Reinforcement Learning (RL), Large Language Models (LLM), and Vision Language Models (VLM). These agents can perform three types of actions: \textbf{(1) Basic Actions:} These are simple movements and interactions using the keyboard and mouse, like moving with ``W'', ``S'', ``A'', and ``D'', attacking with mouse buttons, sneaking with ``E'', dropping items with ``Q'', and using ``Ctrl'', ``Shift'', and ``Space'' for extra moves. \textbf{(2) Mixed Actions:} These actions combine basic actions, like moving back and forth, moving side to side, jumping, sneaking, running, changing the camera angle, and doing things like attacking and using objects. \textbf{(3) High-Level Actions:} These are more advanced, goal-focused actions that make it easier to do things by putting together many basic or mixed actions into one action designed to complete a specific task in the game. By using these action types, Minecraft agents can easily move around and interact with the game world to finish many different tasks.}
\label{minecraft-agent}

\begin{tabular}{ | >{\centering\arraybackslash}m{2.1cm} | >{\centering\arraybackslash}m{2cm} | >{\centering\arraybackslash}m{2cm} | >{\centering\arraybackslash}m{5cm} | }
\hline
\textbf{Model} & \textbf{Method} & \textbf{Action Space} &  \textbf{Task List} \\ \hline  

\textbf{SEIHAI}~\cite{mao2021seihai} & Language + RL&Compound  Action &  Mine diamond\\ \hline

\textbf{VPT}~\cite{vpt} & Vision + RL&Low-Level Action &  Mine log, Craft planks, Craft crafting\_table, Mine cobblestone, Craft stone\_pickaxe, Mine iron\_ore, Craft furnace, Smelt to Iron Ignot, Mine diamond...\\ \hline
\textbf{Steve-1}~\cite{lifshitz2024steve1} & Vision + RL& Low-Level Action & Dig as far as possible, Get dirt, Look at the sky, Break leaves, Chop a tree, Collect Seeds, Break a flower, Go explore, Go swimming, Go underwater, Open inventory, Get dirt, Chop down a tree, Break tall grass... \\ \hline

\textbf{MineDojo}~\cite{fan2022minedojo} & Vision + RL &Compound Action &  Milk cow, Hunt cow, Shear sheep, Hunt Sheep, Combat spider, Combat zombie, Combat pigman, Combat enderman, Find Nether\_portal, Find ocean, Dig hole, Lay carpe...\\ \hline
\textbf{MC Planner}~\cite{mcplanner} &Vision + LLM planning & Compound Action &  Minecraft TASK101 (Craft XXX, Equip XXX, Mine diamond)\\ \hline
\textbf{MC Controller}~\cite{cai2023open} & Vision + RL& Compound Action & Mine oak wood, Hunt sheep, Mine dirt, Mine sand, Mine birch wood, Mine oak\_leaves, Mine birch\_leaves, Obtain wool, Mine grass, Mine poppy,  Combat spider, Hunt wolf, Hunt mushroom cow...\\ \hline

\textbf{Plan4mc}~\cite{plan4mc} &Vision + RL+ LLM planning& Compound Action &  Craft stick, Get crafting table nearby, Craft trapdoor, Craft wooden axe, Craft carpet with shears, Craft hopper with stone pickaxe...\\ \hline
\textbf{Clip4mc}~\cite{ding2023clip4mc} &  Vision + RL& Compound Action &Obtain milk, Obtain wool, Obtain leaf, Obtain sunflower, Hunt cow, Hunt sheep...\\ \hline
\textbf{Ghost}~\cite{zhu2023ghost} & LLM planning& Functional Action & Minecraft Technology Tree (Obtain XXX)\\ \hline
\textbf{Voyager}~\cite{wang2023voyager} & LLM Programming&Functional Action &  Minecraft Technology Tree (Obtain XXX)\\ \hline
\textbf{Steve-Eye}~\cite{zheng2023steveeye} & VLM planning&Not Available &  Craft iron ingot, Find cobblestone, Harvest cobblestone, Find trees, Craft stone axe, Craft and place table, Craft planks, Harvest log...\\ \hline

\textbf{OctoMC(Ours)}& VLM Programming &Functional Action &  Mine a spruce\_log and place it nearby, Mine 3 dirt blocks, Smelt 1 oak\_log, Mine 4 oak\_log and craft 4 oak\_planks and craft 1 craftingtable, Craft 2 chest, Craft 1 oak\_boat, Craft 1 bucket, Craft 1 IronAxe, Mine a jungle\_log and place it nearby, Smelt 1 Chicken, Mine 1 stone and Smelt...\\ \hline
\end{tabular}  
\end{center}  
\end{table}

\newpage %(ls)comment this after editing above paragraph
\subsection{How We Collect OctoMC Training Data}
In the spirit of OctoGibson data collection approach, we've crafted a specialized action space specifically for Minecraft tasks. 

\noindent\textbf{System Message}
\begin{lstlisting}
You are a helpful visual assistant that writes Mineflayer javascript code to complete any Minecraft task specified by me.

Here are some useful programs written with Mineflayer APIs.
I will first give you the name of these programs and then explain how to use them.
await teleport(yaw,distance) //let the bot look at yaw angle and walk with in distance
await mineBlock(bot, name, count) //to collect blocks. Do not use `bot.dig` directly.
await craftItem(bot, name, count) //to craft items. Do not use `bot.craft` or `bot.recipesFor` directly.
await smeltItem(bot, name, "coal" ,count) //to smelt items and using coal as fuel. Do not use `bot.openFurnace` directly.
await placeItem(bot, name, position) //to place blocks. Do not use `bot.placeBlock` directly.
await killMob(bot, name, timeout) //to kill mobs. Do not use `bot.attack` directly.

At each round of conversation, I will give you
Observed Objects:
pic1
yaw=0.00
grass_block(1.22,0.71,3.67)
means the direction of pic1 is yaw=0, and I can perceive grass_block at distance 1.22,0.71 and 3.67
Task Goal: ...
Critique: The direction of next subtask. (If necessary)
Original Subtasks: ...
Previous Action Code: ...
Execution Error: ...
Inventory: ...

You should then respond to me with
Explain (if applicable): Are there any steps missing in your plan? Why does the code not complete the task? What does the chat log and execution error imply?
Plan: How to complete the task step by step. You should pay attention to Inventory since it tells what you have. The task completeness check is also based on your final inventory.
Code:
    1) Write an async function taking the bot as the only argument.
    2) Reuse the above useful programs as much as possible.
		- Use `teleport(yaw,distance)` let the bot look at yaw angle and walk with in distance
        - Use `mineBlock(bot, name, count)` to collect blocks. Do not use `bot.dig` directly.
        - Use `craftItem(bot, name, count)` to craft items. Do not use `bot.craft` or `bot.recipesFor` directly.
        - Use `smeltItem(bot, name, "coal" ,count)` tto smelt items and using coal as fuel. Do not use `bot.openFurnace` directly.
        - Use `placeItem(bot, name, position)` to place blocks. Do not use `bot.placeBlock` directly.
        - Use `killMob(bot, name, timeout)` to kill mobs. Do not use `bot.attack` directly.
    3) Your function will be reused for building more complex functions. Therefore, you should make it generic and reusable.
    4) Functions in the "Code from the last round" section will not be saved or executed. Do not reuse functions listed there.
    5) Anything defined outside a function will be ignored, define all your variables inside your functions.
    6) Call `bot.chat` to show the intermediate progress.
    7) Do not write infinite loops or recursive functions.
    8) Do not use `bot.on` or `bot.once` to register event listeners. You definitely do not need them.
    9) Name your function in a meaningful way (can infer the task from the name).
    10) Try to call teleport to approach the right place before you call other functions.
    11) Each time you should only give me one subtask (not all) with its corresponding code.
    12) You don't need to call the function by yourself.
You should only respond in the format as described below. Besides, I will give you two RESPONSE SAMPLE example for your reference:
RESPONSE FORMAT:
{response_format}  
Explain: ...
Subtasks:
1) ...
2) ...
3) ...
...
Code:
```javascript
// helper functions (only if needed, try to avoid them)
...
// main function after the helper functions
async function yourMainFunctionName(bot) {
	// await teleport(yaw,distance) #plan1: find the sand and teleport
}
```
\end{lstlisting}
\noindent\textbf{Environmental Message}
For each turn, GPT-4 receives information on the two nearest instances of each block type, provided they fall within a maximum range of 20 block units. To manage the context length efficiently, we have adjusted the rotation angle to 60 degrees. This adjustment allows us to generate six snapshots, each accompanied by detailed information about the surrounding blocks. 
\begin{lstlisting}
Observed Objects:
pic1
yaw=0.00
coarse_dirt(6.67,16.36) fern(2.55,5.52) spruce_leaves(4.18,4.18) grass(5.61,5.79) spruce_log(3.67,4.06) poppy(5.7,7.91) large_fern(7.04,7.25) mossy_cobblestone(3.54,3.81) podzol(1.22,1.22) dead_bush(1.58) dandelion(7.58,11.47)
pic2
yaw=1.57
dirt(19.89,20.04) podzol(1.22,1.22) coarse_dirt(6.67,12.98) fern(2.55,2.92) spruce_leaves(4.18,4.18) spruce_log(3.67,4.06) grass(5.61,6.6) dead_bush(7.65,1.58) poppy(8.8,9.67) large_fern(7.04,7.25) brown_mushroom(14.58) mossy_cobblestone(4.95,5.79) dandelion(7.58)
pic3
yaw=3.14
dirt(19.89,20.04) podzol(1.22,1.22) fern(2.55,2.92) coarse_dirt(12.98,14.58) spruce_leaves(4.53,4.53) spruce_log(8.34,9.51) grass(7.52,7.52) dead_bush(7.65,1.58) brown_mushroom(14.58,8.80) mossy_cobblestone(4.53,4.53) large_fern(8.92,10.7)
pic4
yaw=4.71
podzol(1.22,1.22) spruce_leaves(4.53,4.53) grass(5.61,5.79) fern(6.12,6.12) large_fern(8.57,8.69) dandelion(7.58,11.47) spruce_log(8.34,8.8) mossy_cobblestone(3.54,3.81) poppy(5.70,7.91) brown_mushroom(8.80)
Task Goal: Mine a spruce_log and place it nearby
Critique: None

Original Subtasks: None
Previous Action Code: No code
Execution error: No error
Inventory: Empty
Now, please output Explain, Subtasks (revise if necessary), Code that completing the next subtask, according to the instruction above. Remember you should pay attention to the response format and give me just one subtask each turn.
\end{lstlisting}
\noindent\textbf{GPT-4 Response Example}
\begin{lstlisting}
Explain: Since this is the initial request, and the goal is to mine a spruce_log and place it nearby, the first step should be to locate and mine the spruce_log. Based on the observed objects, spruce_logs are visible in multiple pictures but are closest in pic1 (3.67, 4.06). Since inventory is empty, we need to mine at least one spruce_log to complete the task goal.

Subtasks:
1) Teleport near the spruce_log.
2) Mine a spruce_log.
3) Place the mined spruce_log nearby.

Code:
```javascript
async function mineAndPlaceSpruceLog(bot) {
    // Subtask 1: Teleport near the spruce_log
    await bot.chat("Teleporting near the spruce_log...");
    await teleport(0.00, 3.67); // Teleport to the nearest spruce_log based on pic1
}
```
\end{lstlisting}

\section{OctoGTA}
\subsection{Experimental Environment}
OctoGTA is built on top of the renowned video game \textit{Grand Theft Auto V} (GTA V) using SHVDN (Script Hook V .NET), a versatile open-source scripting plugin that enables profound customization and control over in-game elements, transforming GTA V from a gaming platform into a flexible research sandbox.

\subsubsection{Detailed Description of the GTA Environment}
SHVDN allows for the crafting of game scripts using .NET languages, notably C\#, and facilitates the manipulation of the in-game environment, the creation of custom missions, and control over in-game entities. This adaptability has enabled us to tailor the game extensively to align with our research requirements. In the OctoGTA environment, the model is exposed to a myriad of task scenarios and challenges, including walking, swimming, climbing, and engaging in diverse interactions with environmental objects. The abundance of annotated objects within this environment enables the model to interpret its visual inputs more precisely, thereby enhancing learning efficiency.

\subsubsection{Support and Convenience for Model Training}
The GTA environment offers extensive customization options and a range of experimental conditions, such as weather, scenes, and interactive objects, aiding in a comprehensive assessment of the model's performance and adaptability. These features contribute to the anticipated outcomes, which are expected to provide insights and advancements in addressing real-world problems and supporting future research in related fields.

\subsection{Experiment Procedure}
\subsubsection{Task Creation and Setup}
Before the experiment, we prepared the training and test datasets, including a variety of scenes, tasks, and interactive functions, ensuring the model can learn and adapt under diverse conditions. We established 5 different categories of tasks, including having the player get a pet dog into the car, guiding a homeless person to a specific location, assisting in steering the boat towards the coast, and intervening when conflicts occur between pedestrians. For each category, we set five slightly different scenarios, totaling 25 tasks. Upon creation, each task loads the player and the necessary objects and NPCs to the designated locations to complete the task.

\subsubsection{First and Third-Person View Acquisition}
Script Hook V\footnote{\href{http://dev-c.com/gtav/scripthookv/}{Script Hook V} is the library that allows the use of GTA-V script native functions in custom \*.asi plugins.} primarily provides support for native function calls in GTA V's single-player mode, enabling script developers to easily access and set game attributes, coordinates, and other parameters related to characters, interactable items, cameras, and other game elements. We employed SET\_GAMEPLAY\_CAM\_RELATIVE\\\_HEADING from the CAM section and SET\_ENTITY\_HEADING from the ENTITY section for automatic camera rotation, combined with RGB-D image acquisition to automatically gather environmental information.

\subsubsection{Function Construction}
The OctoGTA environment leverages the ScriptHookVDotNet library\footnote{\url{https://github.com/scripthookvdotnet/scripthookvdotnet/}} to construct a comprehensive set of action control functions. These functions are designed to enable the model to interact with the game world and perform a wide range of tasks while maintaining a strong dependence on visual information. A key example of this vision-dependent design is the implementation of the \texttt{goForward(distance)} and \texttt{turnPlayer(degree)} functions. Unlike functions like \texttt{walkTo(location)} that could trivialize the task of reaching a specific location, \texttt{goForward(distance)} and \texttt{turnPlayer(degree)} require the model to actively perceive and navigate the environment. For instance, to reach a desired destination, the model must analyze its surroundings, determine the appropriate direction, and carefully control the player's movement and orientation using these functions. This design ensures that the model's actions are grounded in its visual understanding of the scene, promoting the development of more robust and adaptable embodied AI agents.

In addition to these vision-dependent navigation functions, the OctoGTA environment provides a range of other action control functions for interacting with the game world. These include basic actions such as walking, running, swimming, climbing, and jumping, which allow the player to explore the environment. Furthermore, we have developed functions that facilitate interaction with objects and non-player characters (NPCs) within the scenario, such as entering and driving vehicles, assigning tasks to NPCs, and instructing them to follow or remain stationary.

\subsubsection{Function Generalizability}
One of the key advantages of the action control functions in OctoGTA is their generalizability. The functions are designed to be applicable across a wide range of tasks and scenarios, rather than being limited to specific use cases. This generalizability is achieved through careful function design and parameter selection.

For example, the \texttt{goForward()} function allows the model to control the player's movement in any direction, irrespective of the specific task or location. Similarly, the \texttt{interactWithObject()} function enables interaction with various objects in the game world, regardless of their type or purpose. By providing a consistent interface for interaction, these functions allow the model to learn and apply general strategies for problem-solving and task completion.

The generalizability of the action control functions also facilitates transfer learning and adaptation to novel scenarios. Once the model has learned to utilize these functions effectively in a given set of tasks, it can more easily apply that knowledge to new and unseen situations. This ability to transfer learned skills and strategies is crucial for developing models that can operate in open-ended environments like GTA-V, where the range of possible tasks and challenges is vast and unpredictable.

\subsection{Hand-Crafted Training Data Collection}
Due to the complexity of the GTA-V environment in capturing the environmental messages, we opted for a hand-crafted approach to create the training data for our model. The annotation pipeline is as follows: when the task is initialized, authors who are familiar with the function calls will write functions and plans based on the game screen, and the written functions will be executed in the GTA.
Although time-consuming and labor-intensive, this manual data collection process allowed us to create a high-quality training dataset that is well-suited to the unique challenges of the OctoGTA environment. It is also considered a cold start for future continuous learning based on the OctoGTA environment.

\section{Remarks}
\noindent\textbf{Comparison with EmbodiedGPT:} EmbodiedGPT's core contribution is an embodied-former that cross-attends vision and text to align visual and embodied instructions, while Octopus combines SFT and RLEF.
Although EmbodiedGPT originally did not generate code and uses a frozen LLM, we found that the embodied-former can be applied to any VLM. To fairly compare Octopus, especially the RLEF design, against EmbodiedGPT, we modified Otter by adding an embodied-former (denoted as EmbodiedGPT). Results show that the embodied-former sometimes causes degradation, while RLEF is beneficial, particularly on challenging reasoning and unseen tasks. Nevertheless, we consider the novel problem and environment the primary contribution rather than the inspiring RLEF baseline

\section{Ethical Considerations}
The development of embodied vision-language programming models like Octopus raises several important ethical considerations that need to be carefully addressed as this technology advances.

\noindent\textbf{Responsible Use and Deployment:} Models that can autonomously plan and execute code based on high-level instructions have the potential to be misused if placed in the wrong hands. The developers of such models must implement strict safeguards and guidelines to ensure they are only deployed in responsible and controlled settings by trusted parties. This includes having clear restrictions on the types of tasks the models can be asked to perform. To address this, we will set up a proper license once the code is released.

\noindent\textbf{Safety and Robustness:} In embodied environments, models like Octopus are tasked with taking actions that can have real-world consequences. Extensive testing is needed across diverse scenarios to validate the safety and robustness of the generated plans and code before deployment. Failure cases need to be anticipated with proper exception handling and ``stop'' conditions to prevent harm.